\documentclass[preprint,12pt]{elsarticle}

\usepackage{amssymb}
\usepackage{amsmath}
\usepackage{threeparttable}
\usepackage{tabularx}

\usepackage{tablefootnote}
\usepackage{amssymb}
\usepackage{amsmath}
\usepackage{algorithmic}
\usepackage{algorithm}
\usepackage{threeparttable}
\usepackage{tabularx}
\usepackage{subcaption}
\usepackage{booktabs}
\usepackage{color}
\usepackage{bm}
\usepackage{amsmath}
\usepackage{tablefootnote}

\begin{document}

\begin{frontmatter}



\title{Navigation-GPT: A Robust and Adaptive Framework Utilizing Large Language Models for Navigation Applications}


\author[5,1]{Feng Ma} 
\ead{martin7wind@whut.edu.cn}
\affiliation[5]{organization={State Key Laboratory of Maritime Technology and Safety (SKLMTS), Wuhan University of Technology},
	city={Wuhan},
	country={China}}
\affiliation[1]{organization={Intelligent Transportation Systems Research Center, Wuhan University of Technology},
	city={Wuhan},
	country={China}}

\author[5,1]{Xiu-min Wang}
\ead{349133@whut.edu.cn}

\author[3,6]{Chen Chen\corref{Corresponding}}
\ead{chenchen0120@wit.edu.cn}
\affiliation[3]{organization={School of Computer Science \& Engineering, Wuhan Institute of Technology},
	city={Wuhan},
	country={China}}

\author[4]{Xiao-bin Xu}
\ead{xuxiaobin1980@hdu.edu.cn}	
\affiliation[4]{organization={China-Austria Belt and Road Joint Laboratory on Artificial Intelligence and Advanced Manufacturing, Hangzhou Dianzi University},
	city={Hangzhou},
	country={China}}

\author[5,1]{Xin-ping Yan}
\ead{xpyan@whut.edu.cn}

\affiliation[6]{organization={Nanjing Smart Water Transport Technology Co., Ltd.},
	city={Nanjing},
	country={China}}

\cortext[Corresponding]{Corresponding author}

\begin{abstract}
Existing navigation decision support systems often perform poorly when handling non-predefined navigation scenarios. Leveraging the generalization capabilities of large language model (LLM) in handling unknown scenarios, this research proposes a dual-core framework for LLM applications to address this issue. Firstly, through ReAct-based prompt engineering, a larger LLM core decomposes intricate navigation tasks into manageable sub-tasks, which autonomously invoke corresponding external tools to gather relevant information, using this feedback to mitigate the risk of LLM hallucinations. Subsequently, a fine-tuned and compact LLM core, acting like a first-mate is designed to process such information and unstructured external data, then to generates context-aware recommendations, ultimately delivering lookout insights and navigation hints that adhere to the International Regulations for Preventing Collisions at Sea (COLREGs) and other rules. Extensive experiments demonstrate the proposed framework not only excels in traditional ship collision avoidance tasks but also adapts effectively to unstructured, non-predefined, and unpredictable scenarios. A comparative analysis with DeepSeek-R1, GPT-4o and other SOTA models highlights the efficacy and rationality of the proposed framework. This research bridges the gap between conventional navigation systems and LLMs, offering a framework to enhance safety and operational efficiency across diverse navigation applications.

\end{abstract}




\end{frontmatter}



\section{Introduction}
\label{sec1}
Despite the increasing sophistication of navigation equipment, ships still cannot operate without close supervision of human. Ships’ level of automation and intelligence remains notably lower than that of vehicles, and the accident rate is also difficult to reduce further \cite{ZHANG2025110489,BOLBOT2025110810,ANTAO2023109166}. Investigations reveal that collisions are the primary cause of navigation accidents, with 75\% of collision incidents attributed to erroneous decisions made by crew members \cite{GUO2025110875,WANG2024110201}. Reviewing numerous ship collision incidents, it can be concluded that negligence, unclear understanding of rules, and insufficient consideration of environmental and regulatory factors are the primary causes. Presently, ship navigation still heavily relies on human comprehensive thinking, which imposes a heavy workload. In daily work, ship navigation necessitates the integration of dynamic, multi-source heterogeneous data to achieve a comprehensive understanding of the environment, thereby facilitating accurate decisions. 

Ship navigation and collision avoidance follow established rules, but they also involve numerous uncertainties and contradictions, which are related to regional and meteorological factors. These complexities make it difficult to translate them into traditional computer programs with fixed frameworks \cite{LIU2024107450,LYU2024116530,GLEESON2024119552,SONNTAG2025119907}. In scenarios involving multiple ships, existing navigation systems or expert systems struggle to provide precise and immediate safe routes. Consequently, there is an urgent need for a more effective risk identification mechanism and a flexible framework that considers both safety and cost-effectiveness to address the challenges of daily navigation \cite{ZHU2023116088,DONG2024119512}.

In recent years, the development of artificial intelligence, particularly large language models (LLMs) \cite{radford2019language,raffel2020exploring,liu2020roberta,e2018deep}, has offered new approaches to addressing this challenge. LLMs show potential in dynamic task management with their superior natural language processing and reasoning capabilities \cite{ma2024large,LEE2024100213,NI2024102131}. Research in autonomous driving demonstrates that LLMs possess strong reasoning abilities. These abilities enable LLMs to integrate heterogeneous data and generate context-aware analyses and recommendations, even in unknown scenarios \cite{wen2024dilu,10658060,10491134}. 

Moreover, LLMs can understand driving environments in a human-like manner and address long-tail problems through reasoning, explanation, and memory. Therefore, integrating LLMs into ship navigation and collision avoidance systems is feasible. By efficiently fine-tuning LLMs using data from real-world navigation scenarios, they can better comprehend complex maritime environments and provide intelligent support for ship navigation.

Currently popular LLMs, such as GPT-4, Qwen-2.5, and DeepSeek incorporate techniques like chain-of-thought reasoning and self-correction. Kamoi et al. \cite{tacl_a_00713} have shown that while LLMs can handle certain tasks suited to self-correction, relying solely on the model itself is insufficient for broader self-correction tasks. Furthermore, obtaining real-world feedback is a necessary condition for effective self-correction in LLMs. However, the parameter weights of LLMs are fixed, and they lack the ability to perceive and interact with their external environment. In other words, LLMs do not have sensory capabilities to understand their surroundings or engage with them. This limitation could lead to suboptimal or even dangerous outcomes when LLMs are applied to decision-making tasks. To apply Large Language Models (LLMs) to navigation assistance decision-making, it is necessary to address the dual challenges of how to input various types of external navigation data and how to ensure that the outputted decision support complies with existing navigation rules.

To address the above challenges, this research proposes a dual-core LLM framework, or agent, terms of Navigation-GPT. Firstly, Navigation-GPT leverages a larger LLM core for collecting data and invoking specialized tools. With the assistance of reasoning and action (ReAct) \cite{yao2023react} prompt engineering, the navigation task is decomposed into manageable sub-tasks. This LLM core might invokes tools for subtasks to acquire external information, then collects all navigation-related information and encodes it into the language space, forming structured "prompts", thereby preparing for subsequent decision-making.

Furthermore, task-oriented fine-tuning is performed efficiently using low-rank adaptation (LoRA) \cite{hu2022lora} on the other LLM core. Massive navigation data adhering to COLREGs is utilized, enabling the fine-tuned LoRA module to adapt to the demands of different scenarios and tasks. After LoRA training, this core enhances accuracy and contextual awareness in handling unstructured data, enabling precise lookout scenarios and collision avoidance recommendations in accordance with COLREGs. Experiments conducted across various navigation scenarios have further validated the reliability of dual-core framework, demonstrating the revolutionary potential of LLMs in the field of navigation and collision avoidance.

The main contributions of this research are as follows:

\textbullet \: An LLM is utilized as the control center to establish a novel agent or framework, Navigation-GPT, capable of providing accurate navigation information and collision avoidance recommendations compliant with COLREGs.

\textbullet \: Navigation-GPT integrates ReAct and LoRA techniques, enabling it to make precise navigation decisions based on heterogeneous information from the external environment.

\textbullet \: Systematic experiments validate the reliability and effectiveness of Navigation-GPT in various complex and non-predefined navigation scenarios.

\section{Literature Review}
\subsection{The development of large language models}
In recent years, large language models (LLMs) have made significant advancements in the field of natural language processing (NLP). These models, built on the Transformer architecture, exhibit powerful language understanding and generation capabilities through large-scale parameterization and extensive data training. The Transformer model, first proposed by Vaswani et al. \cite{NIPS2017_3f5ee243}, overcame the limitations of traditional recurrent neural networks (RNNs) \cite{ELMAN1990179} and long short-term memory networks (LSTMs) \cite{6795963} in parallel processing and long-range dependency handling. Its core self-attention mechanism efficiently captures long-distance dependencies within sequences while preserving sequential order through positional encoding, laying the theoretical and technical foundation for subsequent pre-trained language models.

With the widespread adoption of the Transformer architecture, pre-trained language models have become a focal point in NLP research. BERT \cite{devlin-etal-2019-bert} introduced a bidirectional encoder and the masked language modeling (MLM) objective, significantly enhancing language understanding and establishing the pre-training and fine-tuning paradigm. OpenAI released the GPT series, which adopted an autoregressive language modeling approach and demonstrated exceptional performance in text generation tasks through left-to-right sequential generation. Notably, GPT-3, with 175 billion parameters, emerged as the largest language model at the time, showcasing remarkable zero-shot and few-shot learning capabilities. Subsequently, models such as Google PaLM \cite{chowdhery2023palm}, Alibaba Qwen \cite{ZHU2025}, and Meta LLaMA \cite{LI2025110382} demonstrated unprecedented performance in language understanding and generation tasks.

The rapid development of LLMs has revealed emergent abilities \cite{wei2022emergent,10.5555/3600270.3602070}, which refer to the unexpected capabilities models exhibit upon reaching a critical parameter size. These abilities encompass tasks such as language generation \cite{10.5555/3524938.3525989}, complex reasoning \cite{10.5555/3600270.3602070}, multilingual translation \cite{aharoni-etal-2019-massively}, and code generation \cite{feng-etal-2020-codebert}. Contextual learning \cite{lu-etal-2024-emergent} has become a hallmark of LLMs, enabling them to quickly adapt to new tasks with minimal contextual samples, thereby reducing the need for extensive labeled data. Instruction following allows models to execute complex tasks based on explicit natural language instructions, significantly improving capabilities from text generation to multi-turn dialogue generation. Chain-of-thought reasoning, guiding models to perform step-by-step reasoning, enhances decision-making and problem-solving abilities in complex tasks. The emergence of these abilities has expanded the practical applications of LLMs and opened new avenues for exploring model capability mechanisms. Brown et al. \cite{NEURIPS2020_1457c0d6} demonstrated that increasing model size significantly improves task generalization, prompting researchers to investigate the relationship between model size, data diversity, and emergent abilities, offering new theoretical perspectives for model optimization.

In terms of training methods, LLM optimization has undergone several key evolutions. Core methods in the pre-training phase include self-supervised learning, where models learn language patterns and semantic features from unlabeled data by predicting masked or next words \cite{SU2024127063}. This approach avoids reliance on extensive labeled datasets, enabling efficient training on large corpora. To enhance task-specific adaptability, the fine-tuning phase employs supervised learning for parameter updates. Recently, reinforcement learning from human feedback (RLHF) \cite{lang-etal-2024-fine} has become a crucial strategy to improve model performance by optimizing outputs based on human feedback, aligning them better with user expectations. OpenAI developed InstructGPT \cite{NEURIPS2022_b1efde53} using this approach to achieve better user interaction. Lightweight fine-tuning techniques such as LoRA have significantly reduced computational costs, providing greater flexibility for real-world applications. Contrastive learning \cite{cheng-etal-2023-improving}, which maximizes similarities between positive pairs and minimizes those between negative pairs, enhances feature representation capabilities, leading to improved performance in text embedding tasks and a better understanding of complex semantics. Meta-learning further expands the ability of the model to perform well in low-resource environments by learning "how to learn", thereby reducing dependence on large-scale labeled datasets \cite{finn2017model}. Gradient accumulation and clipping techniques have facilitated efficient large-scale model training under memory constraints while preventing gradient explosion, enhancing training stability \cite{NEURIPS2023_8249b30d}.

The latest LLM DeepSeek-R1 represents a groundbreaking milestone \cite{deepseekai2025deepseekr1incentivizingreasoningcapability}. Unlike traditional supervised fine-tuning (SFT) \cite{wei2022finetuned} approaches, DeepSeek-R1 employs large-scale reinforcement learning (RL) through group relative policy optimization (GRPO) \cite{deepseek-math} for base model training, significantly reducing computational costs. Before RL training, the model integrates multi-stage training and cold-start data. Experiments show that DeepSeek-R1, directly fine-tuned with large-scale RL, outperforms state-of-the-art models such as OpenAI-o1-1217 and OpenAI-o1-0912 on benchmarks such as AIME 2024 and MATH-500. On the CodeForces benchmark, it scores only 0.3\% lower than OpenAI-o1-1217. DeepSeek-R1 demonstrates that a purely RL-based approach can significantly enhance the reasoning capabilities of low-density models, showcasing robust self-verification and reflective functions, marking a significant milestone for research in this field.

\subsection{Advanced tasks based on large language models}
LLMs have demonstrated transformative potential across various fields due to their unique capabilities in reasoning, data integration, and context understanding. Notably, in autonomous robotics, LLMs show great promise, particularly in dynamic and safety-critical environments. They can interpret complex sensory information and make environment-aware decisions, advancing autonomous decision-making technologies. Hu et al. \cite{chen2024robogpt} proposed the RoboGPT framework, which optimizes warehouse operations by integrating semantic mapping and adaptive task scheduling, reducing task completion time in dynamic environments. Zhou et al. \cite{10.1007/978-3-031-72667-5_15} developed NavGPT, a navigation assistance system designed for embedded robots. This system utilizes real-time sensor data and multimodal reasoning to achieve efficient obstacle avoidance in unstructured environments. Christos et al. \cite{GKOURNELOS20249} advanced collaborative robotics by designing a framework tailored for team tasks in industrial settings. By incorporating LLMs, the framework enhances communication efficiency and task execution among multiple robots.

In the field of autonomous driving, Liao et al. \cite{LIAO2024100116} proposed the context-aware visual grounding (CAVG) framework, which integrates five core encoders and a multimodal decoder to enhance the ability of autonomous vehicles to interpret the correlation between linguistic commands and visual scenes. Jin et al. \cite{JIN2024105940} developed a government-level framework integrating an LLM to assist urban planners in efficiently evaluating and optimizing parking facilities during the transition period of coexistence between autonomous vehicles and human-driven vehicles.  Alsaif et al. \cite{electronics13244912} focused on industrial equipment, proposing a solution that analyzes multimodal operational data for real-time fault detection and predictive maintenance, crucial for ensuring the long-term stable operation of autonomous vehicles. The system also generates actionable insights to help manufacturers plan maintenance activities in advance, reducing unexpected downtime. 

In summary, LLMs, with human-like reasoning and strong contextual understanding, can perform complex tasks such as autonomous navigation and safety detection. Therefore, integrating LLMs into navigation and collision avoidance is entirely feasible. 

Despite the rapid advancements in the aforementioned general-purpose LLMs, which exhibit strong performance in non-predefined scenarios, significant challenges persist in their application within specialized domains, particularly in the maritime field. Firstly, the maritime domain encompasses a wide range of professional sensors and computational rules that are difficult for general-purpose LLMs to interpret directly. Secondly, the pervasive issue of model hallucinations in LLMs may result in unforeseen and potentially adverse outcomes. Lastly, there is a notable absence of systematic research addressing the integration of rule compliance into LLM outputs, specifically ensuring that decisions adhere to established navigation regulations-a challenge often referred to as the "rule embedding" problem.

\section{Methodology}

\subsection{Overview}
\begin{figure}[H]
	\centering
	\includegraphics[scale=0.27]{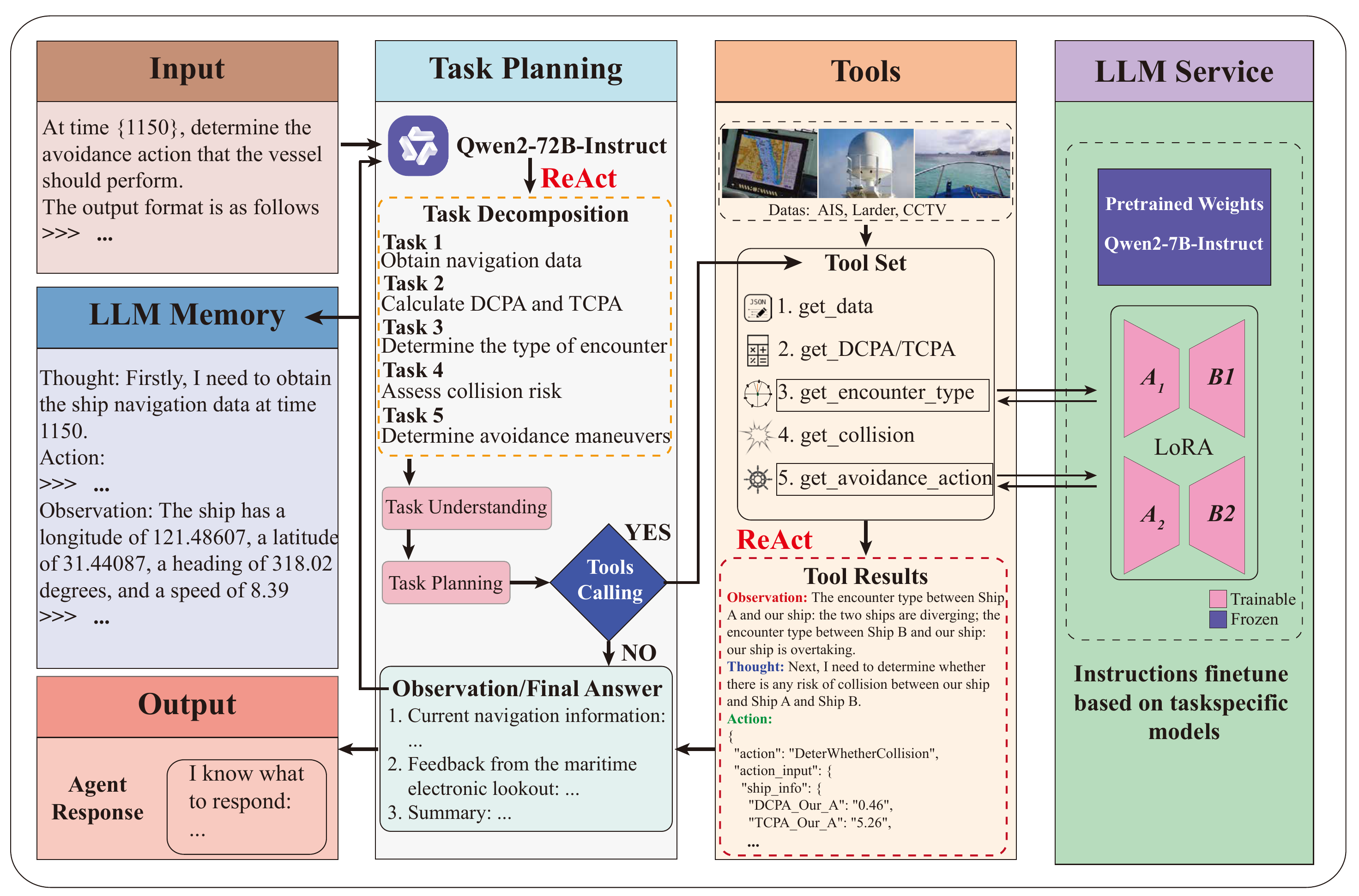}
	\caption{Framework of Navigation-GPT. Navigation-GPT can dynamically plan steps, execute actions, observe outcomes, and adjust subsequent behaviors based on this feedback.}
	\label{Overview}
\end{figure}

In this section, we introduce an agent framework named Navigation-GPT for navigation and collision avoidance tasks. As shown in Figure \ref{Overview}, Navigation-GPT comprises task planning, a memory module, and a set of tools, with a larger LLM core serving as the control center integrating reasoning, action, and memory functions. Tools facilitate interaction between the LLM and the real world. During the reasoning process, this larger LLM core carries out several key operations. It understands and breaks down complex tasks into manageable steps. The framework then invokes tools to interact with the environment and gather necessary information. Heterogeneous data is converted into prompts that the LLM can process effectively. By integrating language analysis with these heterogeneous prompts, the other fine-tuned LLM core provides accurate lookout information to the captain and outputs collision avoidance actions for ships in compliance with COLREGs.

\subsection{Reasoning and Action}
At time $t$, Navigation-GPT receives observations ${o_t} \in O$, executes actions ${a_t} \in \hat A$, and generates a final decision $\pi \left( {{a_t}|{c_t}} \right)$, where ${c_t} = \left( {{o_1},{a_1}, \ldots ,{o_{t - 1}},{a_{t - 1}},{o_t}} \right)$ represents the set of environments and actions.

Since navigation and collision avoidance tasks require complex reasoning to make accurate decisions, Navigation-GPT adopts the ReAct framework. The core principle of ReAct is to expand the action space of LLM to $\hat A = L \cup A$, where $L$ represents the original action space of the LLM, and $A$ represents the set of external tools. ReAct enables the LLM to iteratively generate a trajectory of Thought, Action, and Observation.

Thought belongs to the original action space of the LLM ${\hat a_t} \in L$. It does not interact with the external environment or receive observational feedback. Instead, it involves reasoning based on the current environment to derive useful information, supporting subsequent reasoning and actions. Action represents external tools configured in Navigation-GPT ${\hat a_t} \in A$, which retrieve navigation environment data. Observation ${o_t} \in O$ consists of external information from radar, AIS, and CCTV, fed back to the LLM.
\begin{figure}[H]
	\centering
	\includegraphics[scale=0.195]{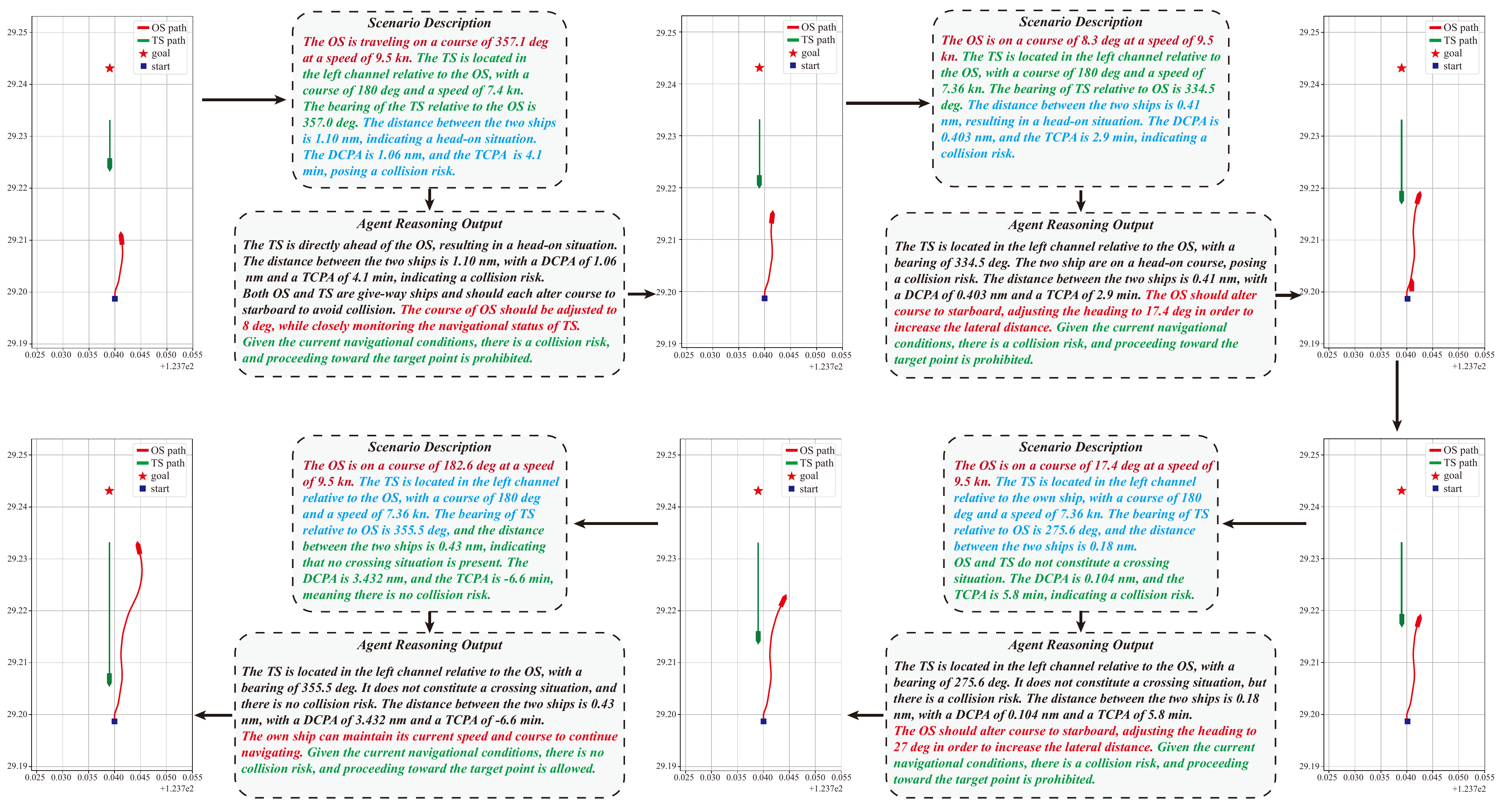}
	\caption{Navigation-GPT generates textual scene descriptions, and the LLM infers collision avoidance actions for the ship.}
	\label{ReasoningandAction}
\end{figure}

As shown in Figure \ref{ReasoningandAction}, to enhance the ability of Navigation-GPT to understand navigation scenarios, the voyage depiction module converts ${o_t}$ into structured text. This module follows linguistic rules to provide standardized natural language descriptions of the ships in navigation. These standardized descriptions include dynamic information about the OS and TS, such as position, speed, direction, and distance.

Navigation-GPT summarizes all behavioral trajectories into the final task decision, avoiding hallucinations that could arise due to the complexity of navigation scenarios. Additionally, the reasoning process refines and adjusts navigation decisions, ensuring better compliance with COLREGs.

\subsection{Tools}
Navigation-GPT integrates five hard-coded tools to perform the following tasks: retrieving onboard sensor data, calculating the distance at closest point of approach (DCPA) and the time to the closest point of approach (TCPA) between two ships, determining the type of ship encounter, assessing collision risk, and executing collision avoidance decisions. Navigation-GPT leverages an LLM to determine the ship encounter type and make collision avoidance decisions.
\begin{figure}[H]
	\centering
	\includegraphics[scale=0.25]{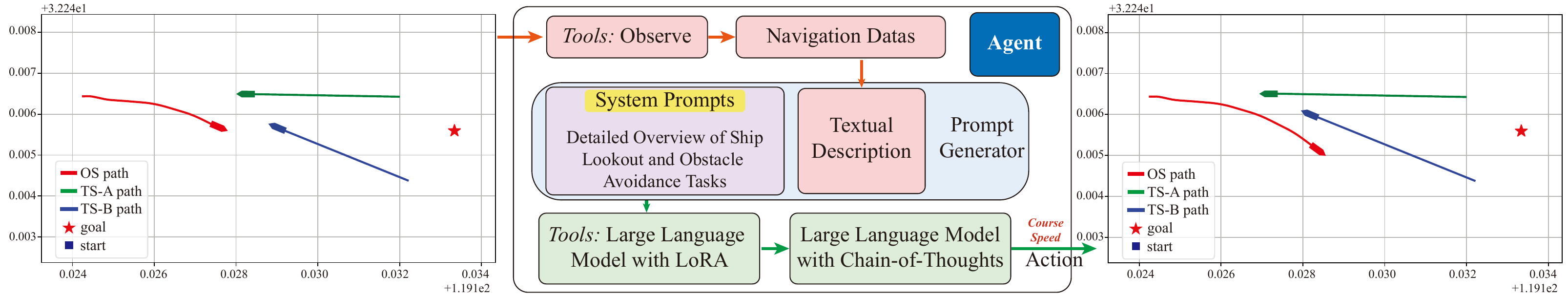}
	\caption{Reasoning and Action. In Navigation-GPT, a smaller-scale LLM with seven billion parameters (Qwen2-7B) and a LoRA module trained for collision avoidance decisions were used to generate collision avoidance strategies. An larger LLM with 72 billion parameters (Qwen2-72B) served as Navigation-GPT's central control system and was integrated with a chain-of-thought mechanism to optimize final navigation decisions.}
	\label{prompts}
\end{figure}

Since a general-purpose LLM cannot accurately understand complex navigation environments, it requires specialized training to adapt to specific tasks. Given the scarcity of accurately labeled navigation data and the diversity of tasks, Navigation-GPT employs LoRA to fine-tune the LLM. LoRA freezes the pretrained weights of the LLM and injects trainable low-rank matrices into selected neural network layers, reducing the number of trainable parameters while enabling rapid task switching. During the fine-tuning process, LoRA constrains matrix updates using the following formula:
\begin{equation}
	\label{p2}
	{W_0} + \Delta W = {W_0} + BA
\end{equation}
where both $A$ and $B$ contain trainable parameters, and where ${{W}_{0}}={{\mathbb{R}}^{d\times k}}$ is the original weight of the LLM, $d$ and $k$ are the dimensions of the model, $r\ll \min (d,k)$ is the rank of a low rank matrix.

As shown in Figure \ref{prompts}, Navigation-GPT analyzes the input task type and invokes the corresponding tools to complete downstream tasks. The prompts provided to the LLM by Navigation-GPT consist of two parts: system prompts and the standardized description of the navigation scenario. The system prompts include a detailed overview of the navigation and collision avoidance tasks, specifying the expected outputs, output format, and constraints for the reasoning process. For each decision, Navigation-GPT concatenates these two prompt components with the feedback from the toolset. The fine-tuned LLM then evaluates and summarizes this information to determine the appropriate collision avoidance actions for the ship.

\section{Experiments}
To evaluate the effectiveness of the proposed Navigation-GPT, this research conducted simulation experiments in various environments. Navigation data from the experimental ship of Wuhan University of Technology was used to construct the simulation scenarios. The research analyzes the collision avoidance decisions of Navigation-GPT in four encounter scenarios, validating the rationality of its decisions. Additionally, the comparison between Navigation-GPT and DeepSeek demonstrates the effectiveness of the technologies within the framework.

\subsection{Data Description and Implementation Details}
This research uses navigation data from coastal water, collected by Wuhan University of Technology in 2024. The dataset includes 500 ships, such as bulk carriers and container ships, covering a total travel distance of 27 kilometers. It consists of real-world navigation scenarios captured using AIS, radar, and CCTV. Based on this data, the ship encounter type dataset (SETD) and the ship collision avoidance decision dataset (SCADD) were constructed.

\begin{figure}[H]
	\centering
	\includegraphics[scale=0.6]{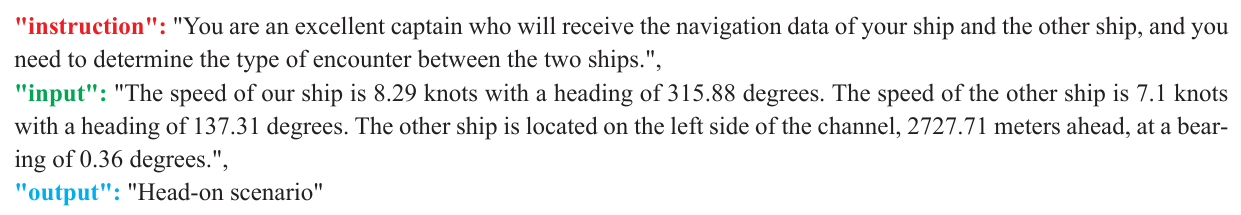}
	\caption{An illustrative example of SETD.}
	\label{SETD}
\end{figure}

\begin{figure}[H]
	\centering
	\includegraphics[scale=0.6]{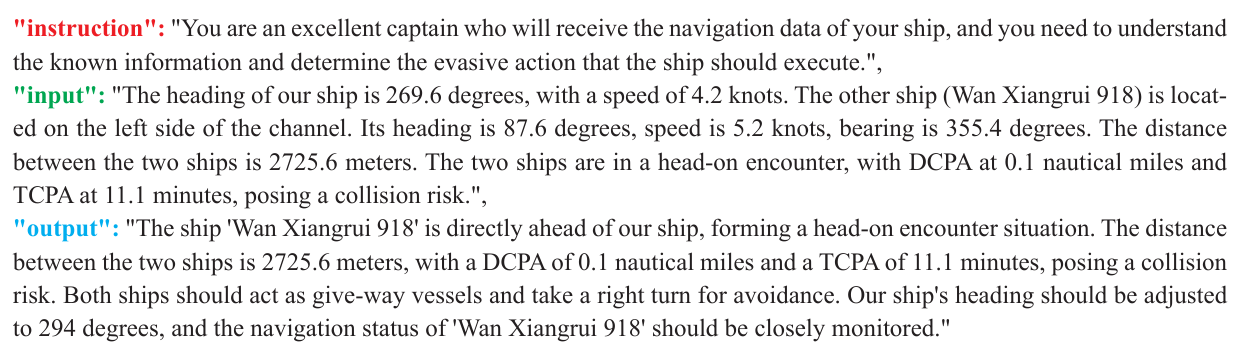}
	\caption{An illustrative example of SCADD.}
	\label{SCADD}
\end{figure}

The SETD consists of 5,000 time-discretized ship encounter scenarios, which describe four types of encounters. These types are classified based on the relative bearing and heading of the TS with respect to the OS, including head-on, small-angle starboard crossing, large-angle starboard crossing, and port crossing. The SCADD contains 150,000 navigation scenarios involving three-ship encounters, with the collision avoidance decisions in these scenarios refined and validated by experienced captains. Figure \ref{SETD} and Figure \ref{SCADD} are examples of SETD and SCADD, respectively

The research presents evaluation results on the test datasets. Additionally, 500 generated prompts were randomly sampled from the SCADD test dataset and assessed for reasonableness by experienced captains. All experiments, or fine-tuning, were conducted on a 4×A100 (40GB) GPU setup. Navigation-GPT employs Qwen2-72B with chain-of-thought reasoning as the central control agent (the larger core) and uses Qwen2-7B as the baseline for LoRA fine-tuning (the smaller decision core). The SETD and SCADD were used for training over 2 epochs, with a batch size of 32, a LoRA rank of 8, a scaling factor of 32, and a learning rate of 1e-4. The LoRA trained on SETD reached the convergence region at 350 steps, with a total training time of 3 hours and 15 minutes. The learning rate was initially set to 1e-4 and gradually decreased to 4.3e-7. The LoRA trained on SCADD reached the convergence region at 1150 steps, with a total training time of 7 hours and 42 minutes. The learning rate started at 1e-4 and eventually decreased to 1.1e-6.

The simulation experiments use a tanker ship model of Fossen \cite{169823}, with its parameters shown in the Table \ref{ship-model}. Additionally, it is assumed that the OS and TS have identical ship handling characteristics. The rudder angle range for the ship model is -30 deg to 30 deg, both for turning and returning.
\begin{table}[h]
	\centering
	\begin{tabular}{l l l l}
		\toprule
		Item & Value & Item & Value\\ 
		\midrule
		Length & 304.8m & Draft &18.46m \\
		Speed & 10kn &  Shaft Speed & 80rpm\\
		Max Rudder Rate & 5deg/s & Max Rudder Rate & 30deg \\
		\bottomrule
	\end{tabular}
	\caption{Parameters of the experimental ship model}\label{ship-model}
\end{table}

\subsection{Case 1: Head-on situations}
\begin{figure}[H]
	\centering
	\includegraphics[scale=0.25]{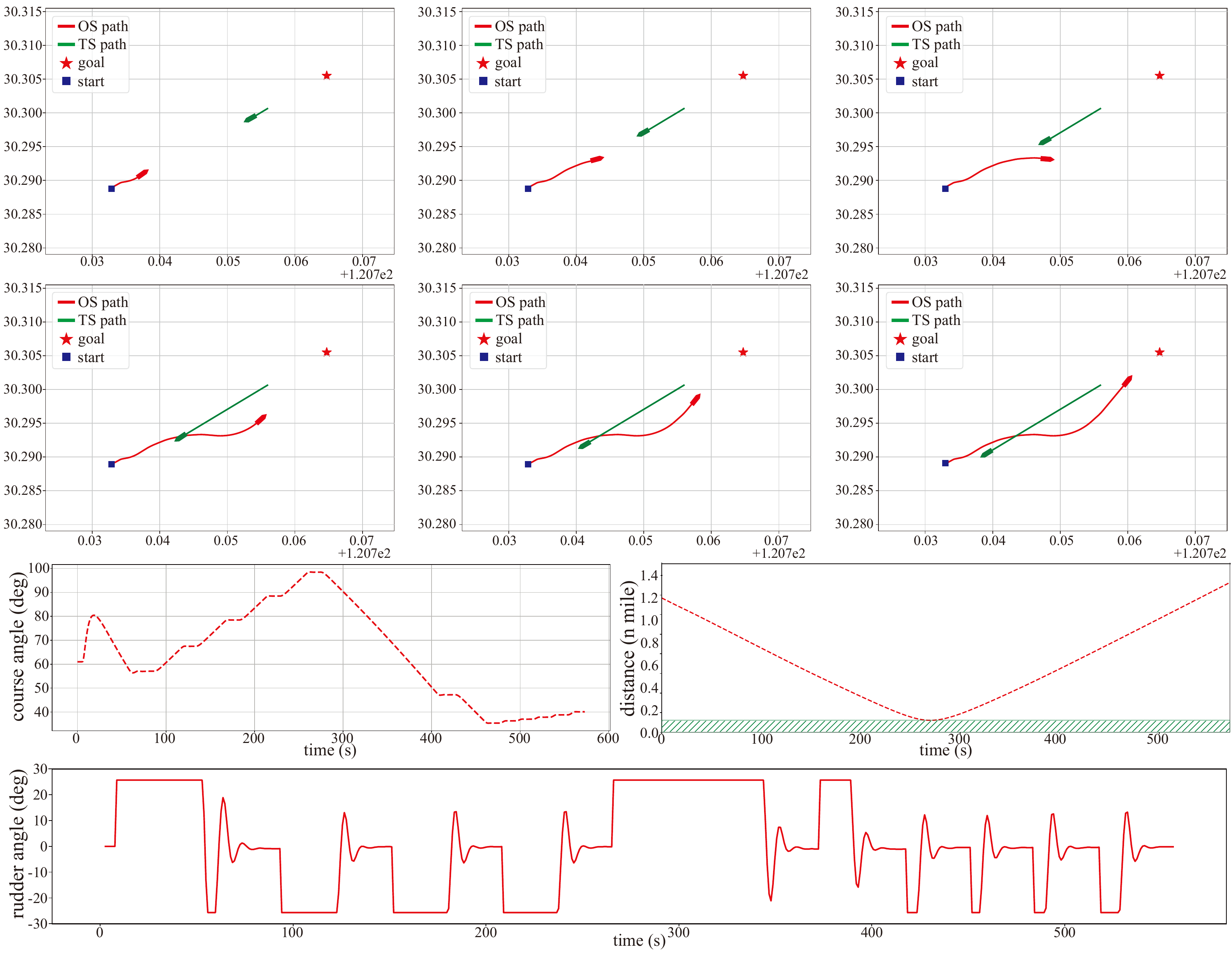}
	\caption{Collision avoidance results for head-on situations, assuming the TS does not alter its course according to COLREGs rule.}
	\label{Head-on}
\end{figure}

As shown in Figure \ref{Head-on}, the TS approaches from the forward side of the OS and maintains its course in accordance with COLREGs. The simulation evaluates the collision avoidance maneuvers of Navigation-GPT under these emergency conditions. According to COLREGs, in a head-on encounter, the OS should turn to starboard and pass the TS on its port side. The risk assessment tool of Navigation-GPT triggers a risk alert, and its obstacle avoidance module retrieves navigation data to propose a collision avoidance decision. At 120 seconds, Navigation-GPT generates and executes this navigation decision. By 280 seconds, the collision is avoided, with the minimum distance between the OS and TS being 0.16 nautical miles. Subsequently, Navigation-GPT identifies the environment as risk-free and advises the OS to resume its course toward the target point.

\begin{figure}[H]
	\centering
	\includegraphics[scale=0.2]{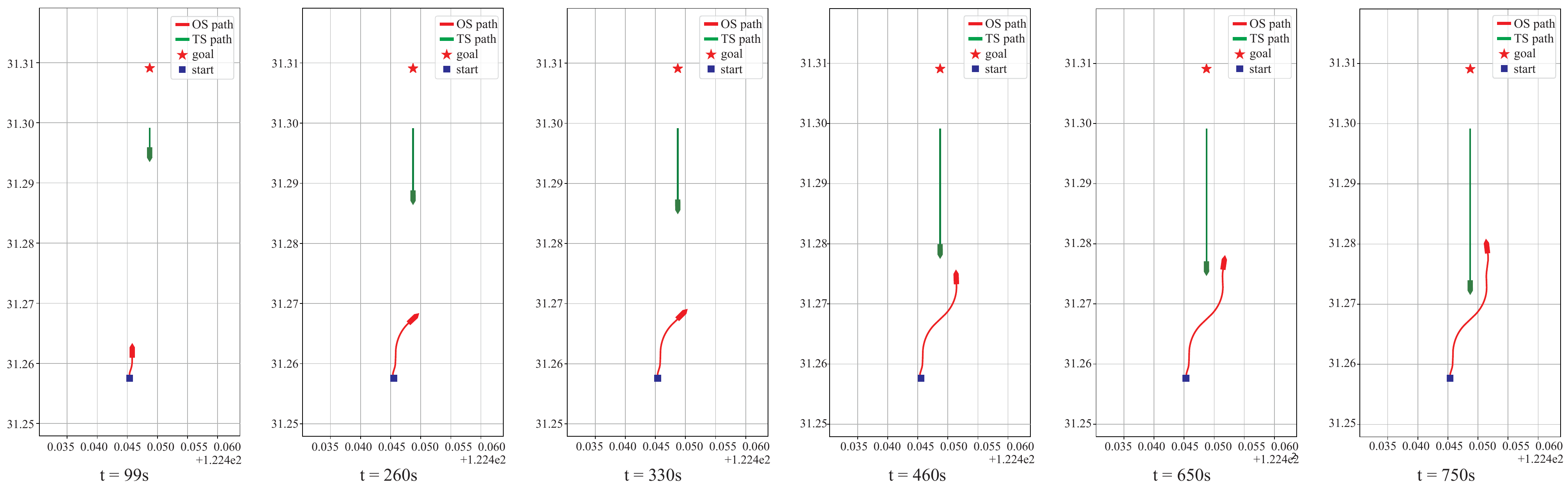}
	\caption{The OS and TS are in a head-on situation. The TS approaches from the starboard side of the OS and maintains its course without altering direction.}
	\label{Head-on-TS-from-right}
\end{figure}
\begin{figure}[H]
	\centering
	\includegraphics[scale=0.2]{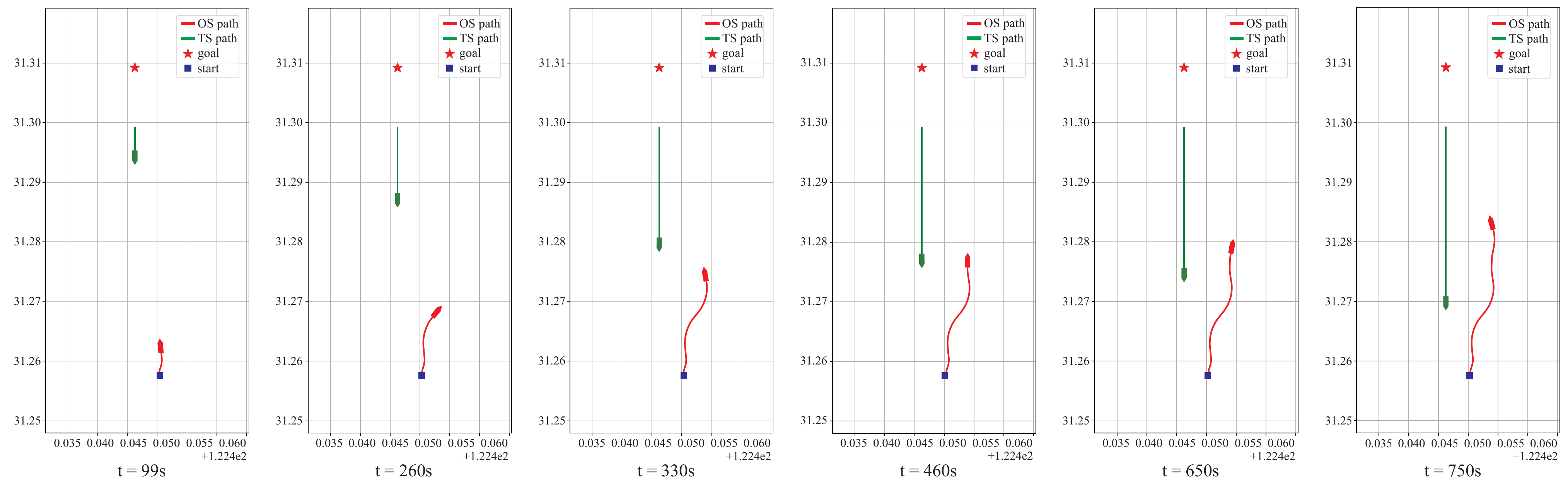}
	\caption{The OS and TS are in a head-on situation. The TS approaches from the port side of the OS and maintains its course without altering direction.}
	\label{Head-on-TS-from-left}
\end{figure}

Figure \ref{Head-on-TS-from-right} and Figure \ref{Head-on-TS-from-left} simulate scenarios where the TS approaches from the starboard and port sides of the OS, respectively, with the TS maintaining its course. As shown in Figure \ref{Head-on-TS-from-right}, the TS is on the starboard side, presenting the greatest challenge. According to COLREGs, the OS is required to turn starboard to avoid a collision. Navigation-GPT accurately identifies the encounter type and performs collision avoidance maneuvers in compliance with COLREGs. At 99s, Navigation-GPT accurately identifies the head-on situation between the OS and TS and begins executing collision avoidance decisions. At 460 seconds, it completes the collision avoidance maneuver in compliance with COLREGs. By 650s, Navigation-GPT adjusts the distance between the OS and TS. Finally, at 750 seconds, Navigation-GPT determines that the navigation environment is free of collision risks and resumes heading toward the target point.

In such scenarios, the navigation strategy of Navigation-GPT is both efficient and cautious, and its text output can be directly used as a decision-making reference for the operator.

\subsection{Case 2: Crossing situations}
\begin{figure}[H]
	\centering
	\includegraphics[scale=0.145]{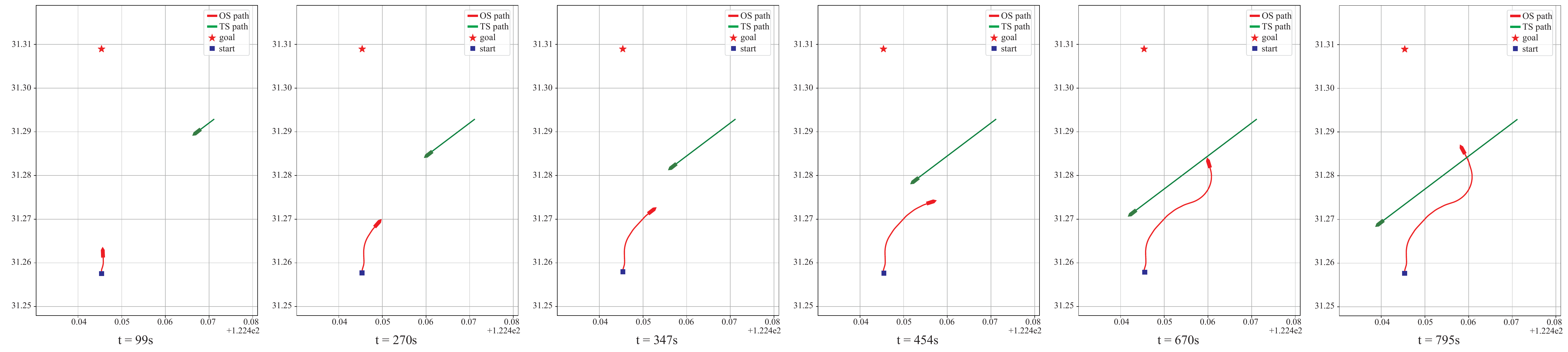}
	\caption{The OS and TS are in a crossing situation with a small angle on the starboard side. The TS approaches from the starboard side of the OS and maintains its course without altering direction.}
	\label{Cross-TS-from-small}
\end{figure}

\begin{figure}[H]
	\centering
	\includegraphics[scale=0.125]{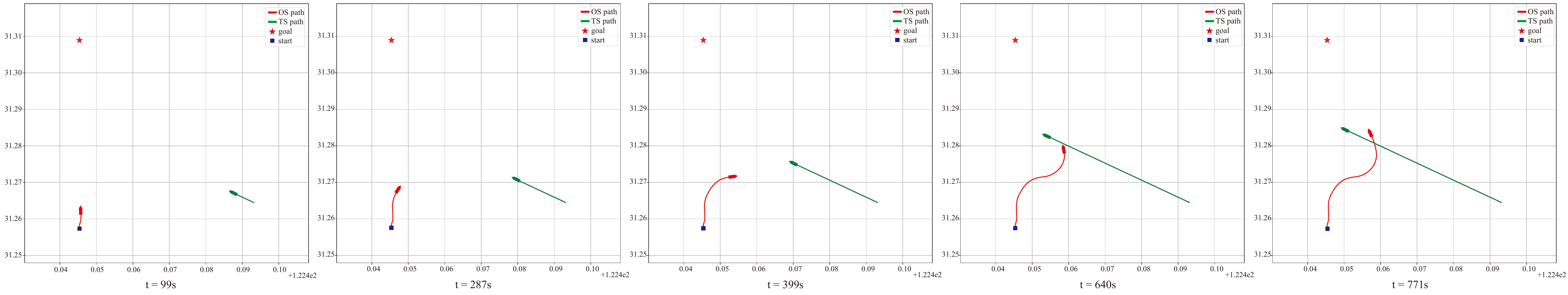}
	\caption{The OS and TS are in a crossing situation with a large angle on the starboard side. The TS approaches from the starboard side of the OS and maintains its course without altering direction.}
	\label{Cross-TS-from-large}
\end{figure}

In crossing scenarios, this research designs three TS with different navigation directions to test Navigation-GPT compliance with COLREGs. As shown in Figure \ref{Cross-TS-from-small}, the TS approaches from the starboard side of OS, maintaining constant speed and course with a small relative bearing. Navigation-GPT identifies this as a small-angle crossing on the starboard side. According to COLREGs, TS is the stand-on ship, while OS is the give-way ship and should turn to starboard to avoid TS. At 454 seconds, OS completes the avoidance maneuver and, after confirming no collision risks in the current navigation environment, resumes course toward the target point.
As shown in Figure \ref{Cross-TS-from-large}, TS approaches from the starboard side of OS with a larger relative bearing. Navigation-GPT identifies risk at 287 seconds and initiates a more significant starboard turn. OS successfully avoids TS by 640s.

\begin{figure}[H]
	\centering
	\includegraphics[scale=0.19]{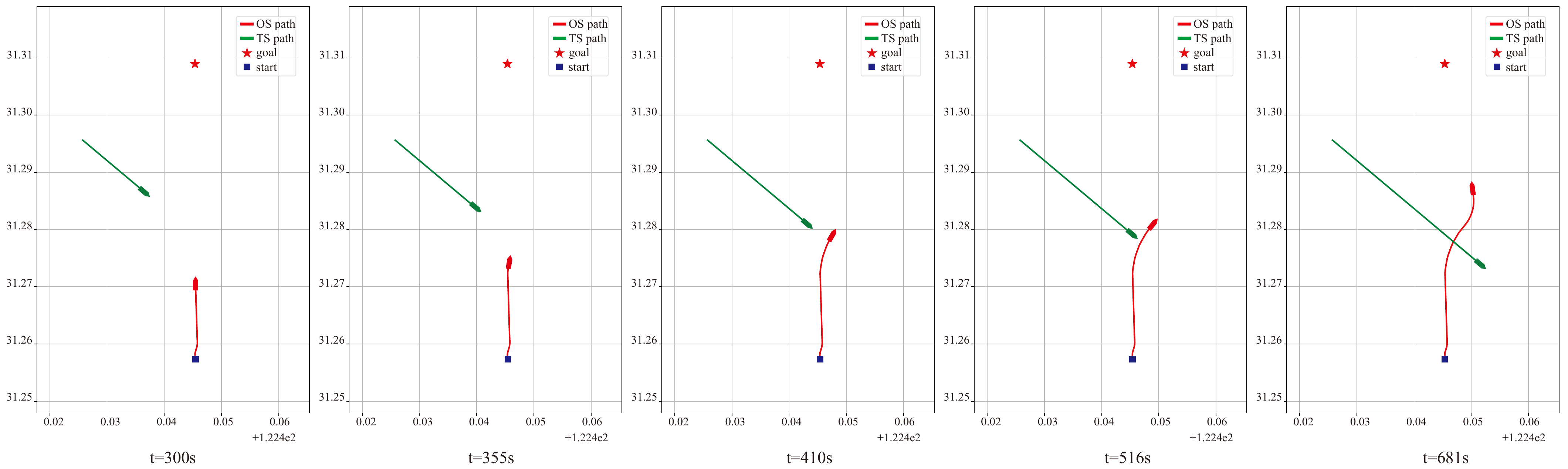}
	\caption{The OS and TS are in a crossing situation. The TS approaches from the port side of the OS and maintains its course without altering direction.}
	\label{Cross-TS-from-left}
\end{figure}

As shown in Figure \ref{Cross-TS-from-left}, the TS approaches from the port side of OS, forming a port-side crossing. According to COLREGs, OS is the stand-on ship, while TS is the give-way ship. Before 300 seconds, OS maintains speed and course toward the target point. At 355 seconds, Navigation-GPT determines that current navigation conditions cannot be resolved by the actions of a single ship. Therefore, Navigation-GPT advises OS to proactively avoid collision by turning to starboard to increase distance from TS. At 516 seconds, OS completes the avoidance maneuver and resumes course toward the target point.

\begin{table}[h]
	\centering
	\resizebox{\textwidth}{!}{
	\begin{tabular}{l l l l l l}
		\toprule
		Parameters & Value & Parameters & Value & Parameters & Value\\ 
		\midrule
		Start of OS & (122.445374, 31.257936) &  Start of TS A  & (122.481637, 31.285263)&Start of TS B& (122.469149, 31.294028)\\
		
		Goal of OS & (122.445374, 31.307936) & Goal of TS A   &(122.429876, 31.276136) & Goal of TS B &(122.441838, 31.278260)   \\
		
		Speed of OS & 8.0 knots  & Speed of TS A  & 10 knots  &Speed of TS B& 6 knots \\
		
		Course of OS & 0 degree & Course of TS A   & 260 degrees &Course of TS B    & 240 degrees \\
		\bottomrule
	\end{tabular}}
	\caption{Initial Parameters for the Three-Ship Encounter Scenario}\label{rightrightcrossThree-ShipEncounter}
\end{table}

\begin{figure}[H]
	\centering
	\includegraphics[scale=0.145]{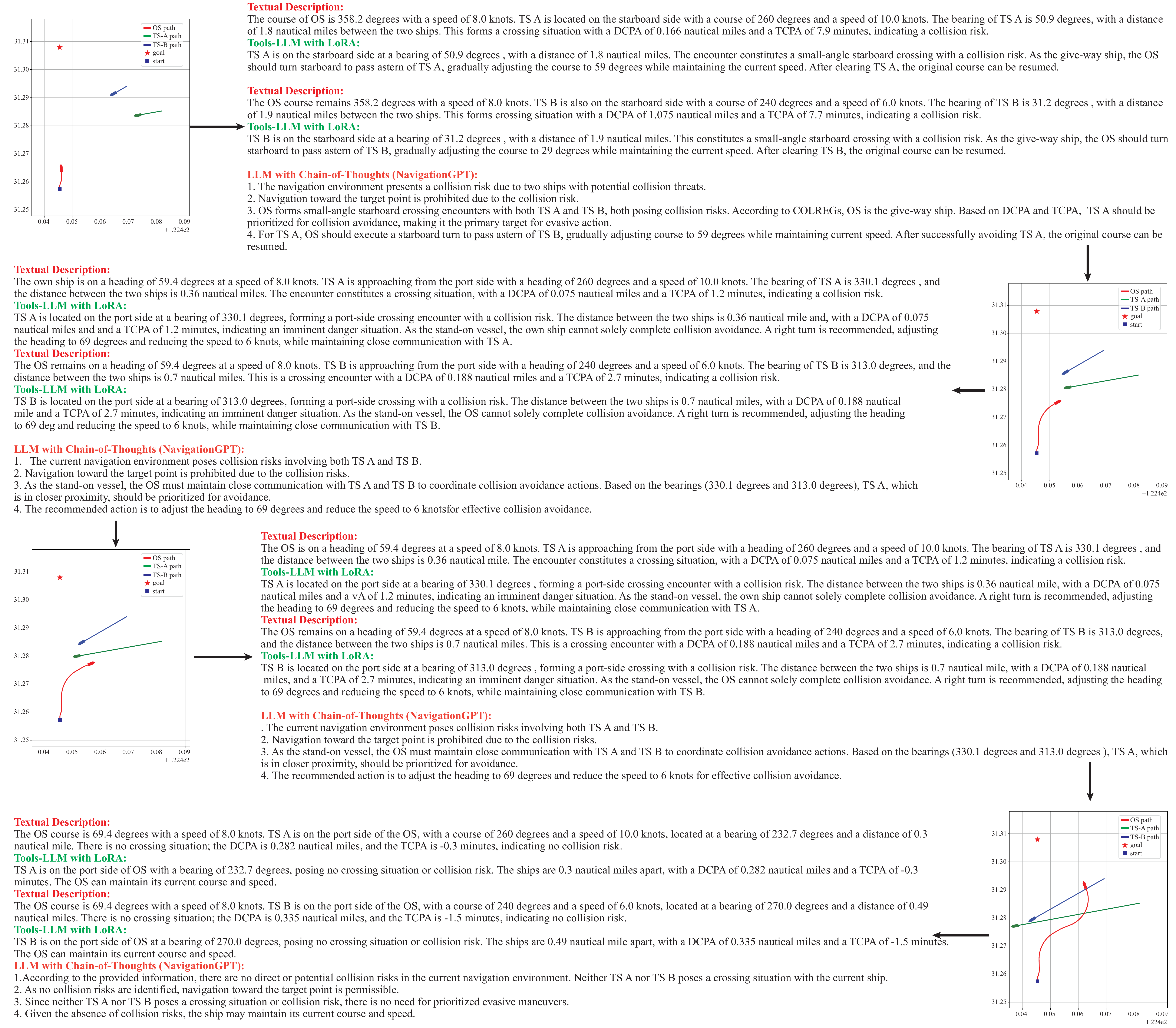}
	\caption{Navigation-GPT reasoning and ship trajectory changes in a three-ship crossing scenario.}
	\label{Cross-TSTS-prompts}
\end{figure}

\begin{figure}[H]
	\centering
	\includegraphics[scale=0.33]{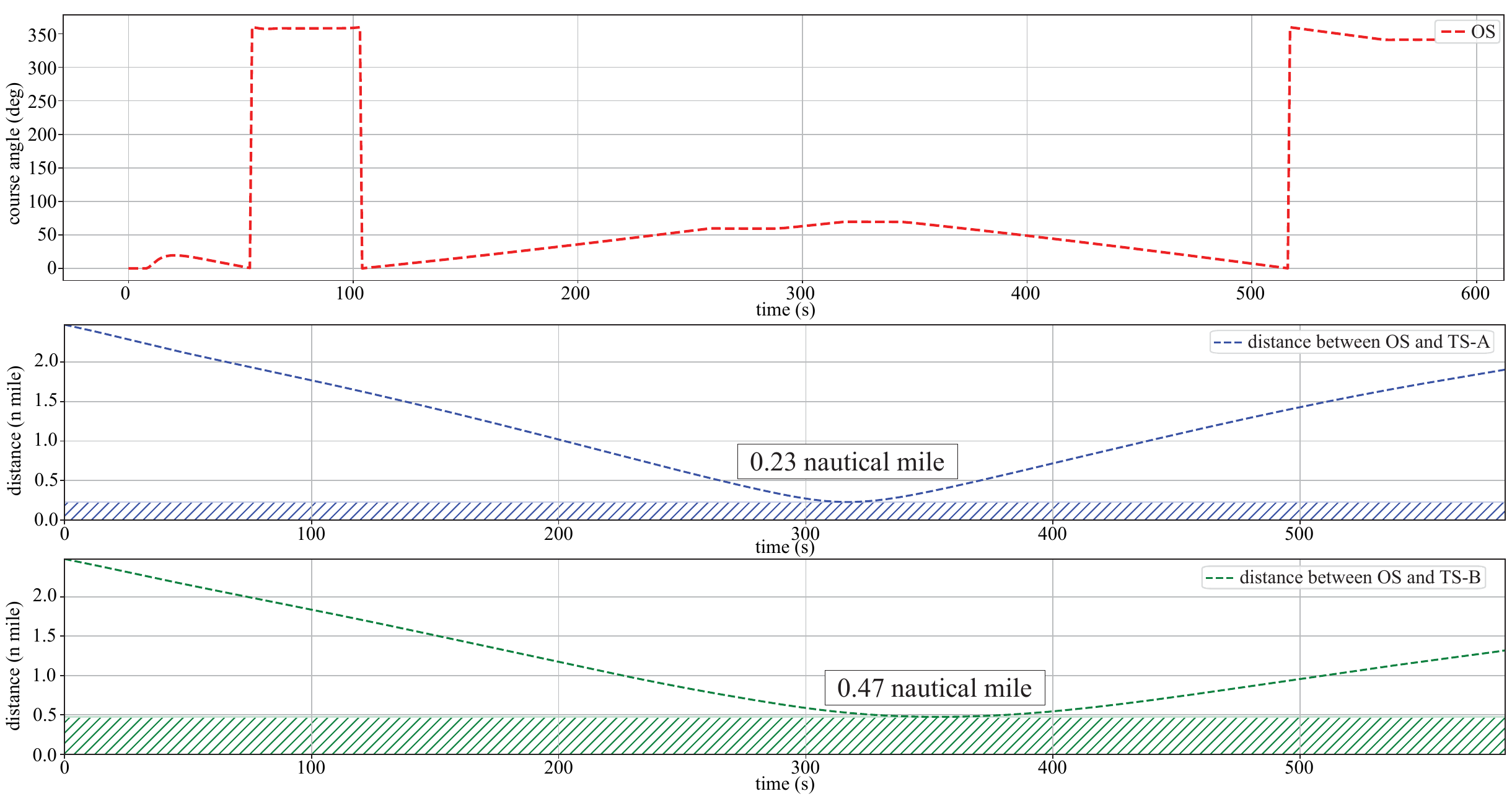}
	\caption{Heading and distance changes of OS relative to TSs in a three-ship crossing scenario.}
	\label{Cross-TSTS-courge}
\end{figure}

Additionally, this research employs Navigation-GPT to provide collision avoidance decisions in a three-ship crossing scenario, validating its effectiveness. Table \ref{rightrightcrossThree-ShipEncounter} shows the initialization parameters of three ships. As shown in Figure \ref{Cross-TSTS-prompts}, Navigation-GPT accurately identifies the encounter types involving two TS and generates appropriate collision avoidance strategies. Subsequently, the embedded chain-of-thought LLM further evaluates the priority of ships for avoidance based on DCPA, TCPA, and distance information, deriving collision avoidance decisions suitable for the current navigation scenario. Navigation-GPT provides feedback to the operator on the current navigational environment, fulfilling the lookout function, and updates the ship model with the latest speed and heading to complete the collision avoidance task. Figure \ref{Cross-TSTS-courge} illustrates the changes in OS heading. Throughout the avoidance process, OS maintained a safe distance from both TS A and TS B.

In more complex encounter scenarios involving three or more ships, the proposed framework can still exhibit decision-making capabilities comparable to human performance, and its performance is already on par with commonly studied optimization methods.

\subsection{Case 3: Being Overtaking situations}
\begin{figure}[H]
	\centering
	\includegraphics[scale=0.24]{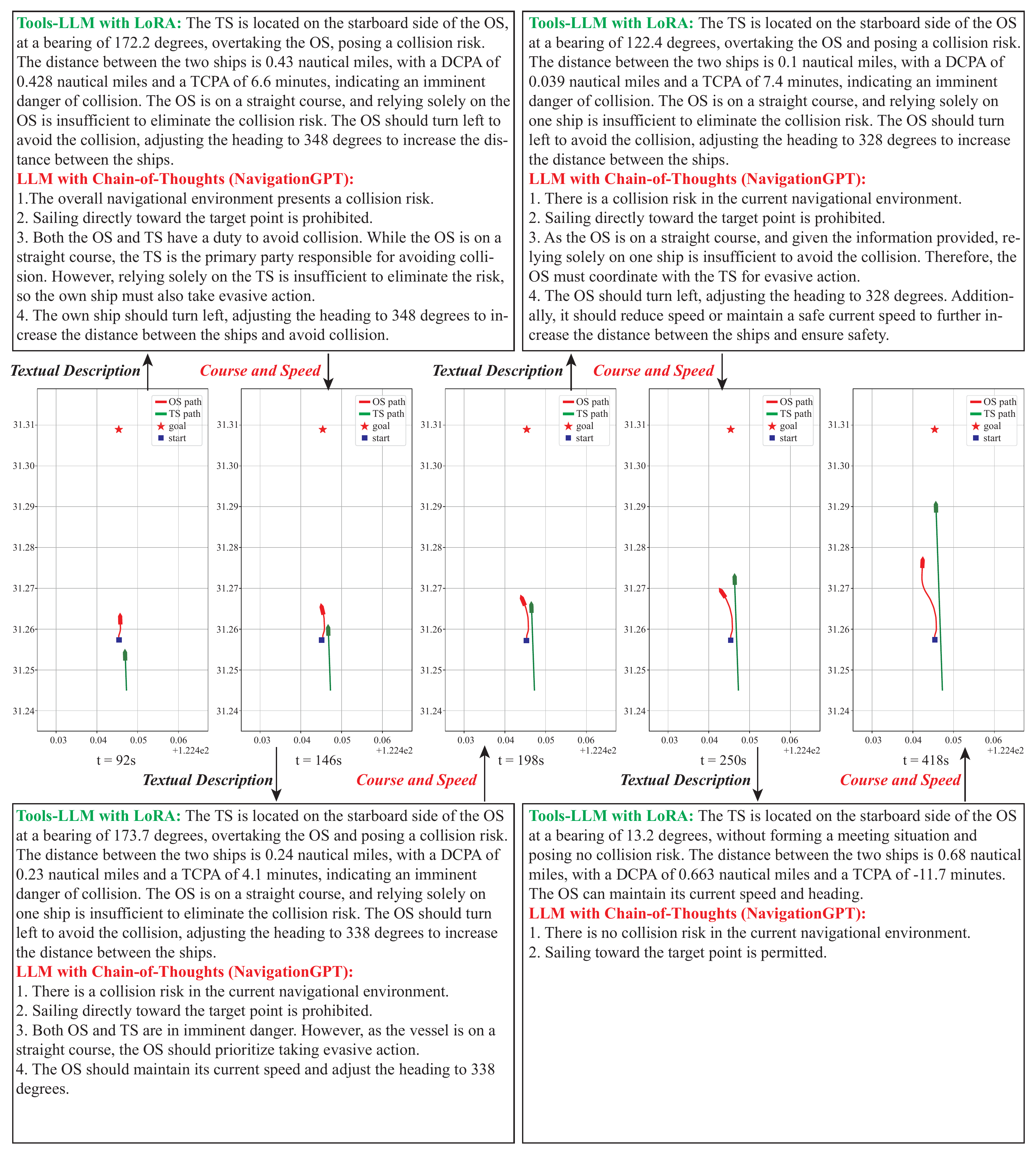}
	\caption{The reasoning of Navigation-GPT and the changes in ship trajectories is analyzed. TS serves as the overtaking ship (give-way ship) in the overtaking scenario but does not adjust its course and speed in accordance with COLREGs regulations.}
	\label{Overtaking-os2ts}
\end{figure}

As shown in Figure \ref{Overtaking-os2ts}, Navigation-GPT identifies that the TS approaches from the right rear of the OS, overtaking it. According to COLREGs, the TS turns right to overtake. However, in the simulation, the heading and speed of TS remain constant. Thus, before 92 seconds, the OS maintains a direct course toward the target point. At 92 seconds, Navigation-GPT detects that the distance between the OS and TS is below the safe threshold, and the TS, maintaining constant heading and speed, rapidly closes in. The OS actively turns left to increase the distance between the two ships and avoid a collision. From 92 seconds to 250 seconds, the OS heading changes from 348 degrees to 328 degrees, successfully avoiding the TS. By 250 seconds, the TS completes the overtaking maneuver. Based on DCPA, TCPA, and the distance between the ships, Navigation-GPT determines that no collision risk exists in the current navigational environment and recommends proceeding toward the target. At 418 seconds, the heading adjusts to 13.2 degrees.

During the process of being overtaken by another ship, the proposed framework maintains its own stability and adopts robust navigational decisions when faced with potential threats.

\subsection{Case 4: Unexpected or Non-predefined Scenarios}
In the Case 1, 2, 3, traditional computer programming, decision tree models, and Bayesian network models could also be implemented, and might even be more efficient, since these scenarios are relatively predictable. The core issue preventing existing computer decision systems from being applied to ship navigation lies in the undefined, sudden scenarios, which are particularly common on ships. 

Large language models provide an intelligent agent that mimics human thinking and can make relatively reasonable decisions in undefined or un-predictable scenarios. This advantage is particularly evident in human conversations, where questions from people are always varied, un-predictable, yet well-designed LLMs can always respond flawlessly. The source of this generalization capability has not yet been fully explained by foundational research. 

\begin{figure}[H]
	\centering
	\includegraphics[scale=0.13455]{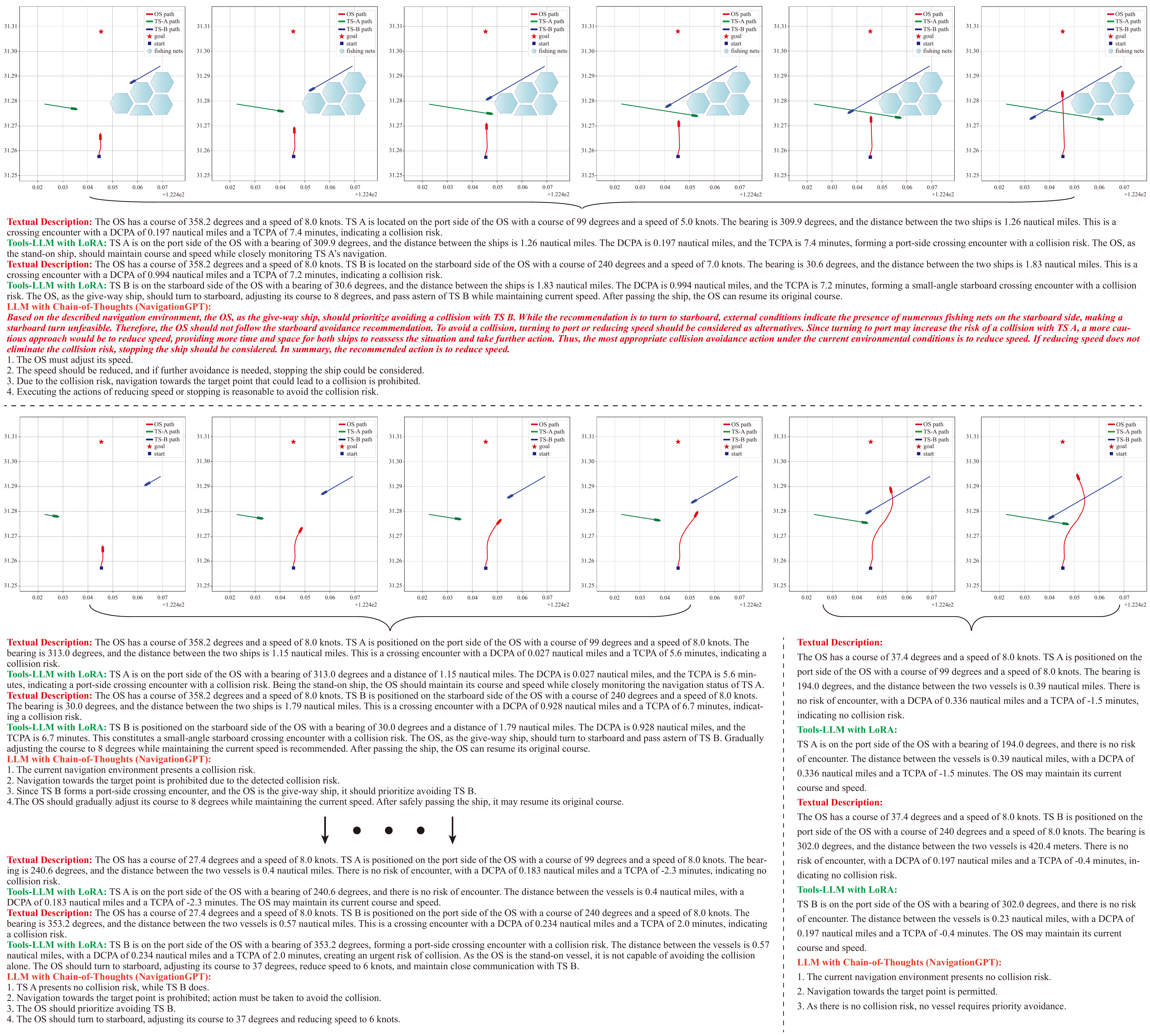}
	\caption{The reasoning process for Navigation-GPT in providing navigation recommendations when there is a sudden change in the channel. TSs maintain course and speed unchanged.}
	\label{fishing-nets}
\end{figure}

In Case 4, this research incorporates the navigation environment prompt "a large area of fishing nets suddenly appears on the starboard side" into the reasoning process of Navigation-GPT. As shown in Figure \ref{fishing-nets}, Navigation-GPT converts navigation data into text and obtains collision avoidance recommendations from LoRA. LoRA suggests that the OS should turn to starboard according to COLREGs to avoid TS B. However, upon detecting an unexpected appearance of extensive fishing nets on the starboard side, Navigation-GPT determines that a starboard turn would negatively impact the OS. Consequently, it recommends abandoning the turn and completing collision avoidance by reducing speed or stopping. Notably, the developers did not write any pre-programmed branches instructing the intelligent agent to slow down to avoid risks. This solution was entirely "thought up" by Navigation-GPT in a dilemma, demonstrating its ability to solve unknown challenges. This capability is crucial in ship navigation. The lower part of Figure \ref{fishing-nets} depicts a scenario in normal waters, where Navigation-GPT accurately follows COLREGs to avoid collisions. After confirming a risk-free environment, it advises the OS to navigate toward the target point.

\begin{figure}[H]
	\centering
	\includegraphics[scale=0.16]{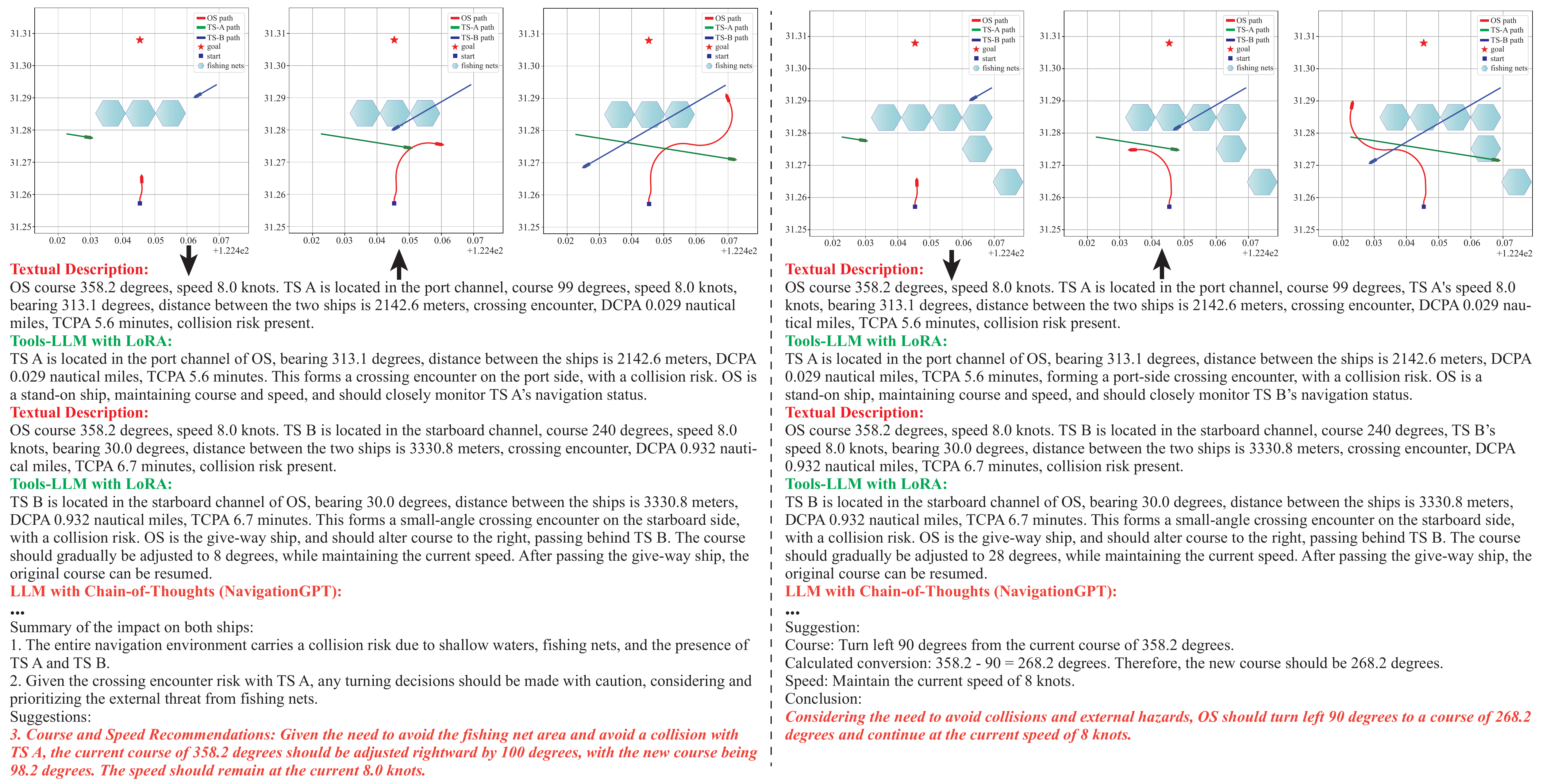}
	\caption{The reasoning process for Navigation-GPT in providing navigation recommendations when there is a sudden change in the channel. A large area of fishing nets suddenly appears ahead and to the right of OS. TSs maintain their course and speed unchanged.}
	\label{fishing-nets-head-right}
\end{figure}

As shown in Figure \ref{fishing-nets-head-right}, when the sensor detects a large fishing net suddenly appearing ahead of the OS, LoRA provides a normal collision avoidance decision to the agent. However, Navigation-GPT identifies that the suggested route cannot safely navigate out of the hazardous area. Consequently, Navigation-GPT modifies the recommendation from LoRA with a more aggressive heading adjustment. When fishing nets are detected ahead and to the right of the OS, Navigation-GPT disregards the suggestion from LoRA and proposes a left turn to avoid the collision and safely exit the dangerous waters.
\begin{figure}[H]
	\centering
	\includegraphics[scale=0.16]{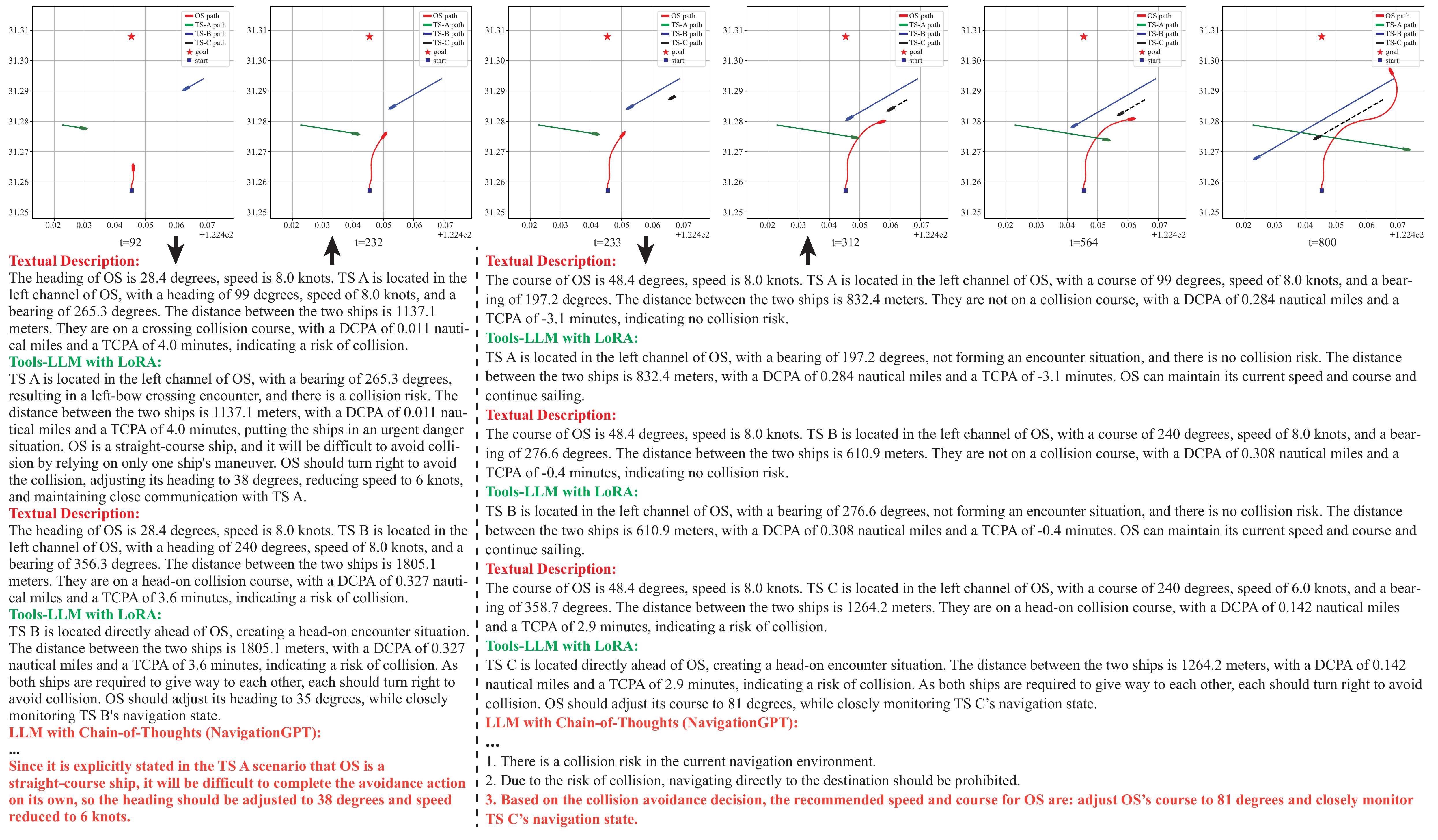}
	\caption{The reasoning process for Navigation-GPT in providing navigation recommendations when there is a sudden change in the channel. An unplanned ship suddenly appeared on the radar. TSs maintain their course and speed unchanged.}
	\label{unexpect-ships}
\end{figure}

As shown in Figure \ref{unexpect-ships}, the sensor feedback indicates only two TSs in the navigation data, and Navigation-GPT conducts collision avoidance planning for them. However, at 233 seconds, an unplanned ship is suddenly detected by the sensor. Despite the OS already being in a three-ship collision avoidance state, Navigation-GPT quickly recalculates the avoidance plan based on the latest data in this sudden scenario. Navigation-GPT accurately identifies the meeting type between OS and TS C, while also considering the other TSs, ultimately providing safe navigation recommendations and successfully completing the collision avoidance task for the unexpected ship.

The experiments designed additional unexpected scenarios, but due to space limitations, these are not individually described. The tests demonstrate that this companion-like intelligent agent maintains strong decision-making stability when responding to unknown and sudden situations. Extensive preliminary LoRA training also effectively instilled a form of "risk aversion" in the finalized Navigation-GPT. Its decision-making tends to be conservative and compliant, aligning with the original design intent.

\subsection{Case 5: A Comparative Research with Deepseek and GPT-4o}
To evaluate the collision avoidance capabilities of Navigation-GPT, this research conducted a comparative analysis with the DeepSeek-R1-Distill-Qwen-32B (DeepSeek) model in a three-ship encounter scenario. 

\begin{figure}[H]
	\centering
	\includegraphics[scale=0.186]{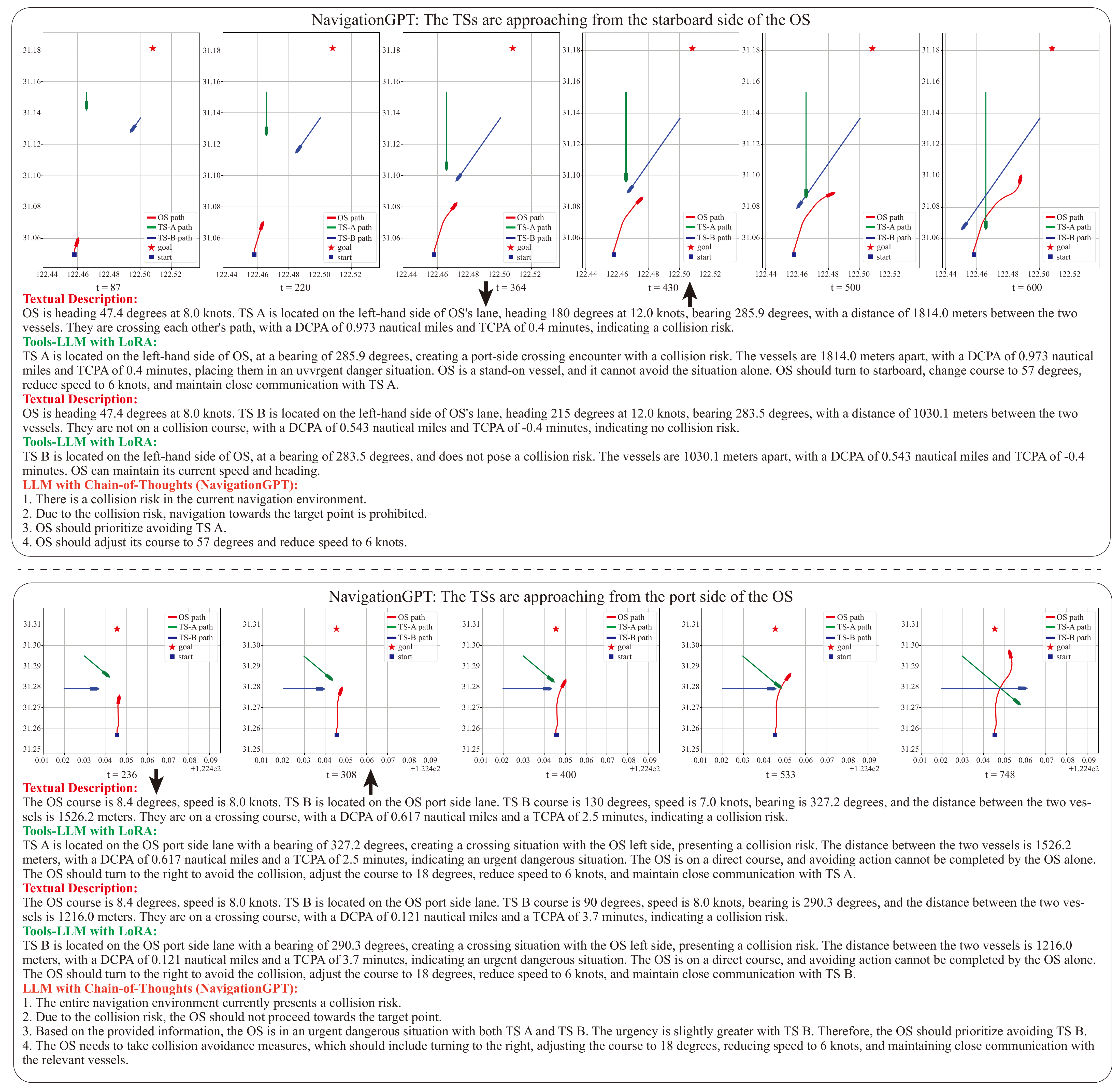}
	\caption{Navigation-GPT in a three-ship encounter scenario makes collision avoidance decisions. The speed and heading of TS A and TS B remain unchanged.}
	\label{NavigationGPT-success}
\end{figure}

As shown in Figure \ref{NavigationGPT-success}, Navigation-GPT detects a potential collision risk between TS A, TS B, and OS at the 87 seconds. The LoRA module of Navigation-GPT outputs collision avoidance suggestions for the TSs, which are then analyzed by the larger LLM core, leading to the final avoidance action for the ship. From the 220 seconds to the 500 seconds, OS gradually adjusts its course and successfully avoids the collision. At the 600 seconds, Navigation-GPT detects that the navigation environment poses no collision risk and recommends that OS proceed towards the target point.

\begin{figure}[H]
	\centering
	\includegraphics[scale=0.186]{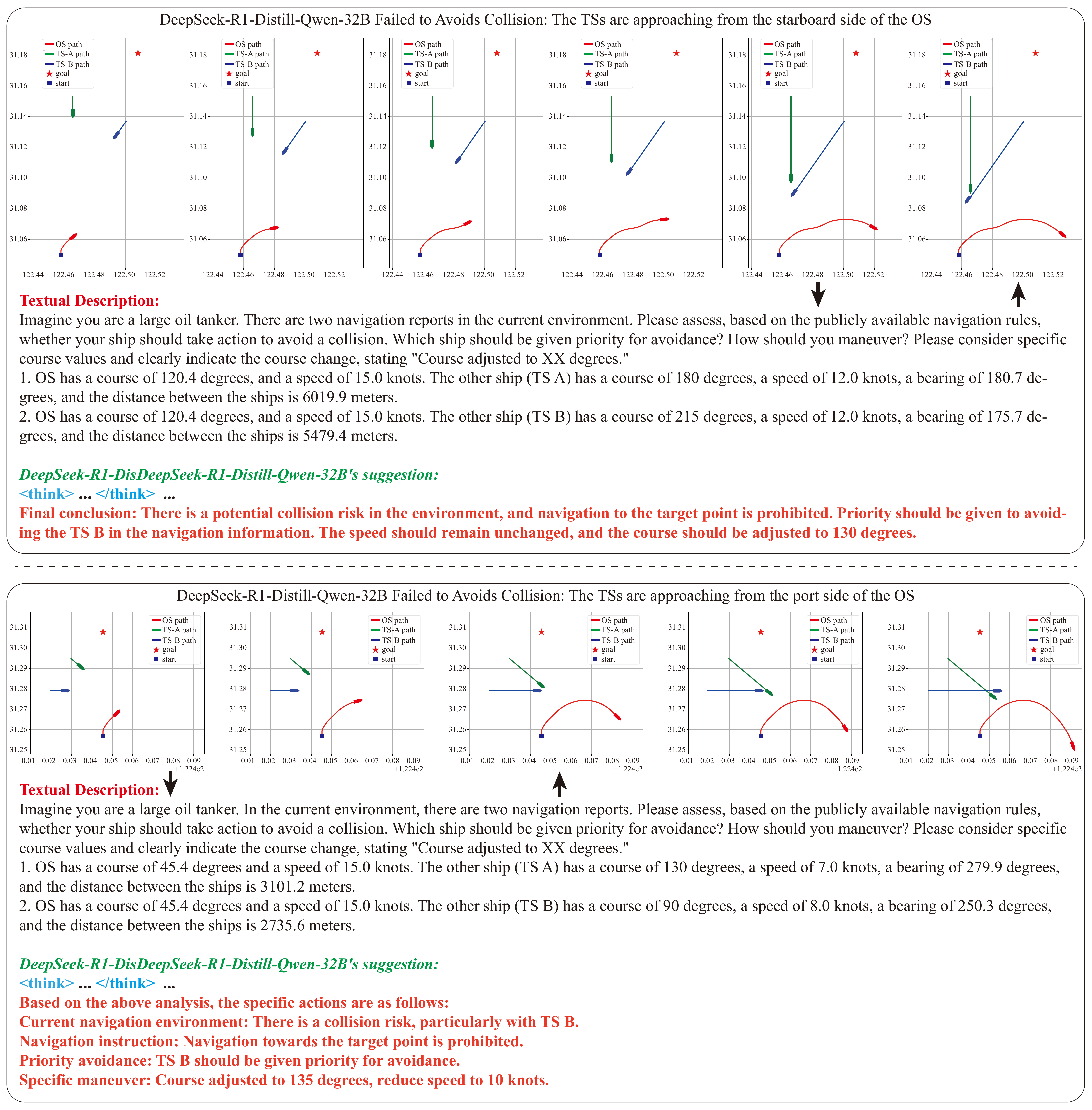}
	\caption{DeepSeek in a three-ship encounter scenario makes collision avoidance decisions. The speed and heading of TS A and TS B remain unchanged.}
	\label{fail-deepseekright-left}
\end{figure}

As shown in Figure \ref{fail-deepseekright-left}, without the integration of external tools, DeepSeek begins to execute collision avoidance tasks upon receiving basic ship information. Regardless of whether TSs are located on the port or starboard side of the OS, DeepSeek is initially able to proactively perform avoidance maneuvers based on the COLREGs. However, after completing the avoidance maneuver, DeepSeek struggles to accurately interpret basic environmental information. Furthermore, there are deviations in the risk parameters, such as the DCPA, calculated by DeepSeek, which is a key reason for the model performing excessive collision avoidance actions. Experimental results indicate that while DeepSeek possesses some spatial understanding capabilities, its computational limitations hinder the full realization of this potential.

\begin{figure}[H]
	\centering
	\includegraphics[scale=0.185]{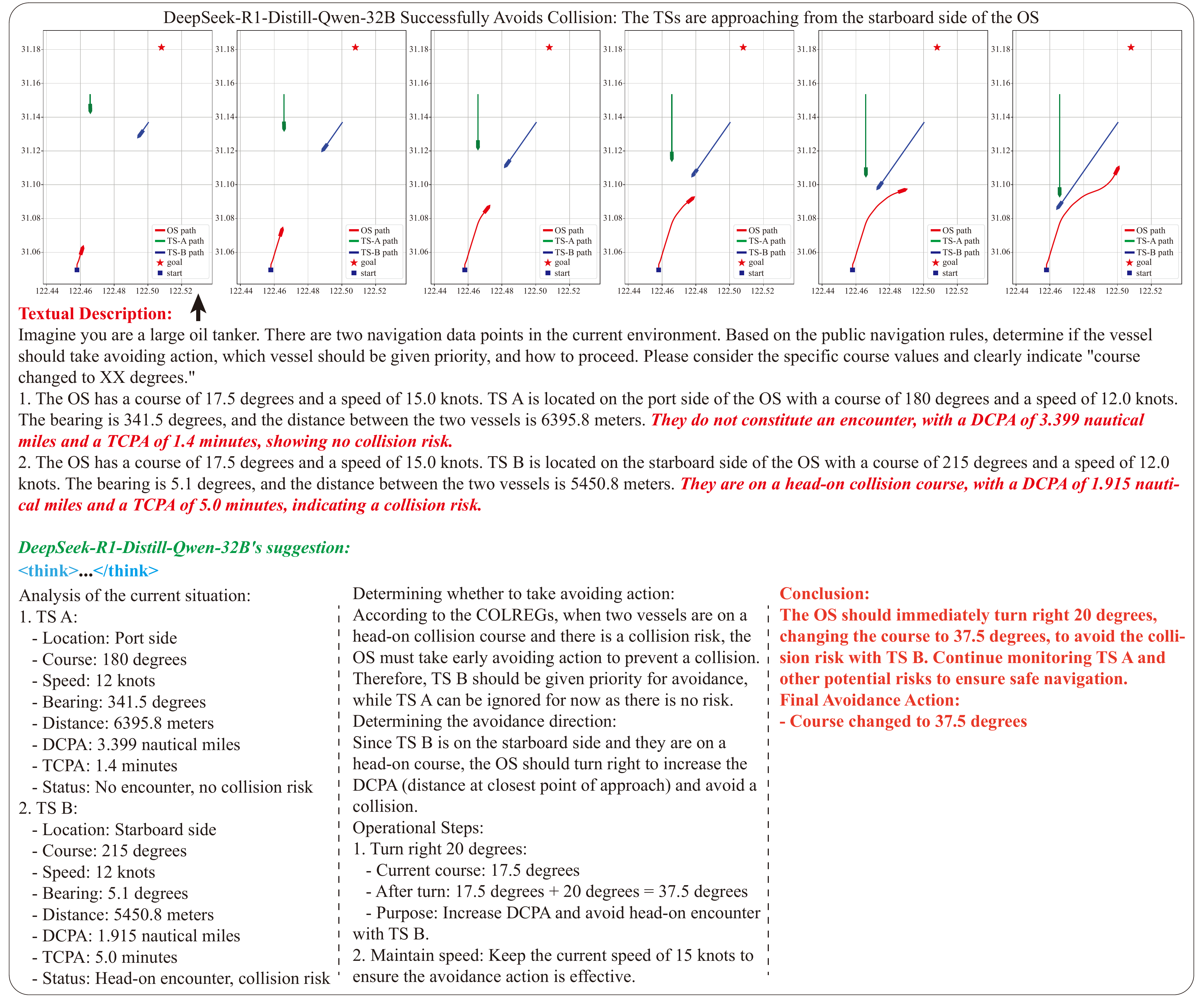}
	\caption{DeepSeek successfully avoids the TSs in the three-ship encounter scenario. The TSs are approaching from the starboard side of the OS. DeepSeek receives the risk parameters calculated by the toolset from Navigation-GPT. The speed and heading of TS A and TS B remain unchanged.}
	\label{deepseek-success-right}
\end{figure}
\begin{figure}[H]
	\centering
	\includegraphics[scale=0.185]{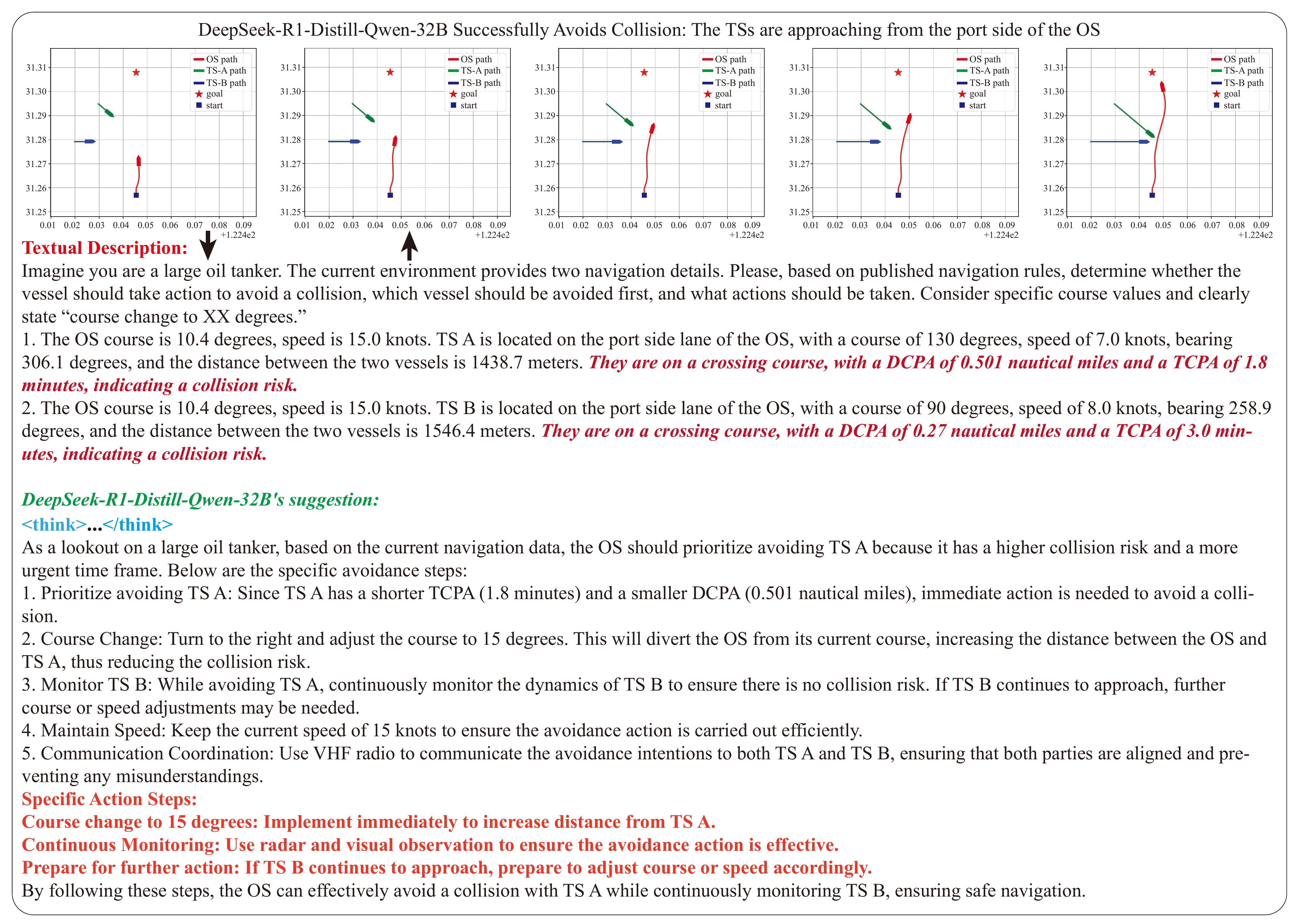}
	\caption{DeepSeek successfully avoids the TSs in the three-ship encounter scenario. The TSs are approaching from the port side of the OS. DeepSeek receives the risk parameters calculated by the toolset from Navigation-GPT. The speed and heading of TS A and TS B remain unchanged.}
	\label{deepseek-success-left}
\end{figure}
\begin{figure}[H]
	\centering
	\includegraphics[scale=0.122]{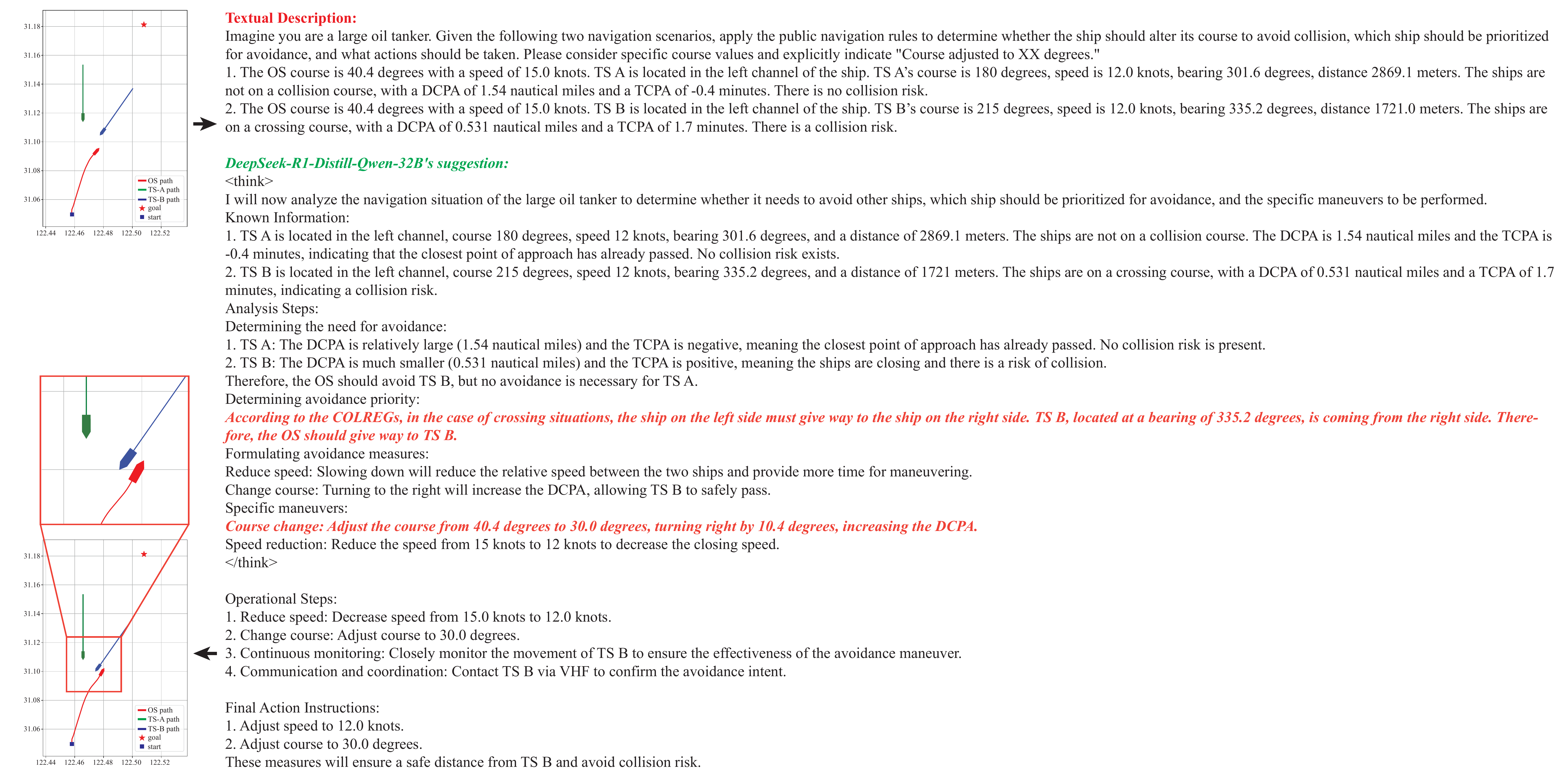}
	\caption{Even with the integration of external tools, DeepSeek still exhibits hallucinations. The scenario description explicitly states that TS B is located on the port side of the OS. However, DeepSeek incorrectly assumes that TS B is on the starboard side during the inference process.}
	\label{hallucinations}
\end{figure}

As shown in Figure \ref{deepseek-success-right} and Figure \ref{deepseek-success-left}, after integrating the toolset from Navigation-GPT, DeepSeek plays the role of the decision-making core of Navigation-GPT. The new framework avoids the complex collision parameter calculations performed by the model itself. The scenario description includes accurate data feedback from tools such as DCPA and TCPA. Using the more comprehensive navigation information, DeepSeek accurately avoids TS A and TS B. After OS passes the TSs, DeepSeek prevents unnecessary avoidance maneuvers and successfully proceeds towards the target point. When serving as the decision-making LLM core, DeepSeek can play a role similar to the fine-tuned small core of Navigation-GPT. However, its output decision structure is inherently unstable, and its wording often does not conform to navigational norms. Additionally, the excessively long chain-of-thought (CoT) process results in lower operational efficiency compared to the small core of Navigation-GPT. The DeepSeek used 58 seconds to generate a decision in average, while the proposed method only needs 15 seconds. Moreover, as shown in Figure \ref{hallucinations}, DeepSeek without specialized fine-tuning can still experience model hallucinations when serving as the decision-making core.

\begin{figure}[H]
	\centering
	\includegraphics[scale=0.185]{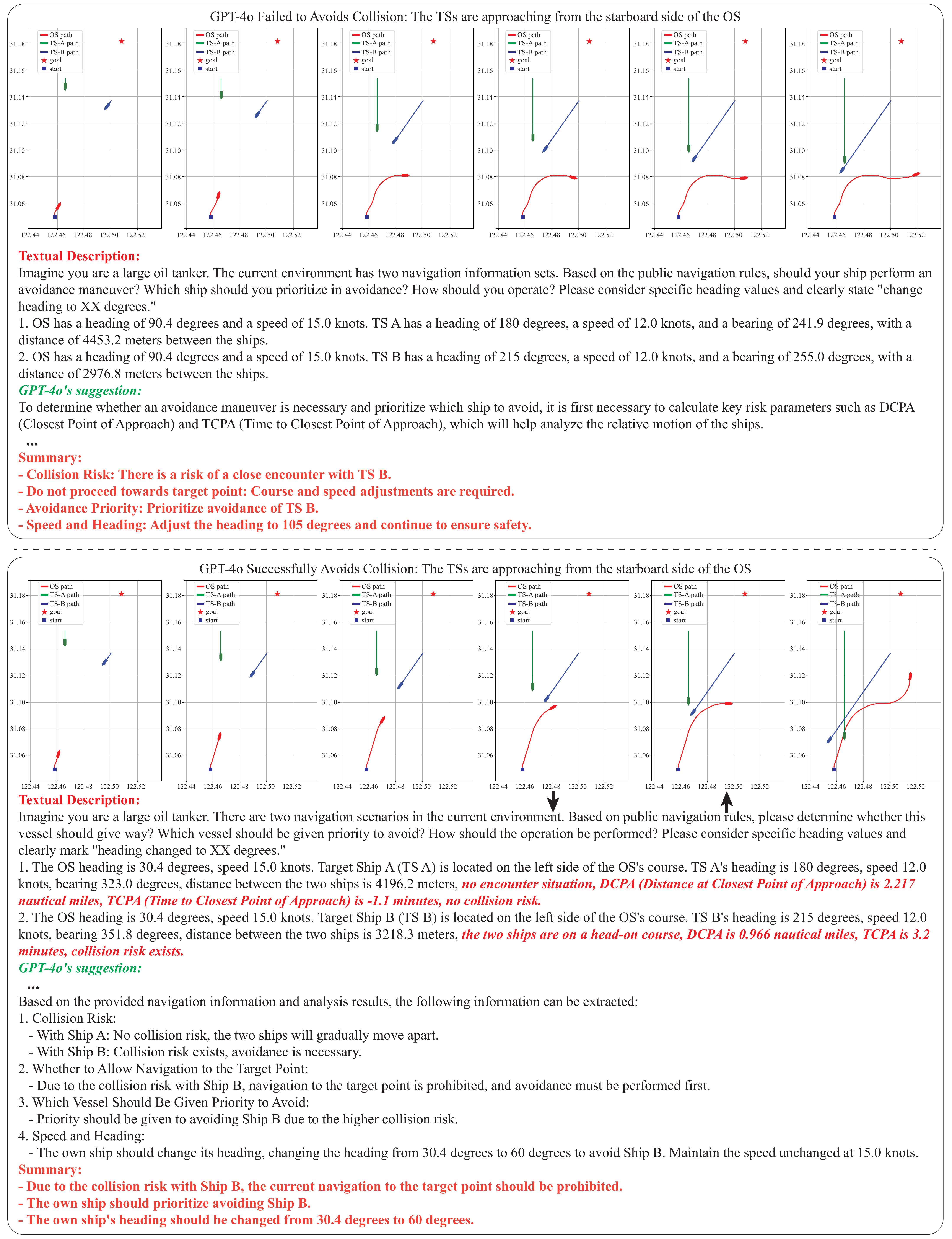}
	\caption{The collision avoidance process of GPT-4o in a three-ship encounter scenario. The TSs approach from the right side of the OS. In the upper part, GPT-4o is not connected to external tools. In the lower part, GPT-4o is connected to external tools. The speed and course of TS A and TS B remain unchanged.}
	\label{gpt4o-success-fail}
\end{figure}

As shown in Figure \ref{gpt4o-success-fail}, similar to DeepSeek, when GPT-4o is not connected to external tools, it cannot accurately assess the risk in the encounter scenario. Therefore, after the OS completes collision avoidance, GPT-4o provides unnecessary avoidance suggestions. After connecting to external tools, the OS trajectory stabilizes. Upon completing collision avoidance, GPT-4o ultimately recommends that the OS navigate toward the target point.

When GPT-4o is used as the decision-making core, similar to DeepSeek, although its performance improved, it occasionally exhibited hallucinations. Moreover, its navigation decisions are more aggressive compared to both DeepSeek and the method proposed in this research. As an online model that cannot be deployed locally, GPT-4o is, in fact, very challenging to implement on board ships.

\section{Discussion}
For many years, the maritime industry has endeavored to design obstacle avoidance or watchkeeping procedures that comply with navigation rules to assist seafarers. This objective appears to be relatively straightforward to achieve, as various navigation rules and conventions can typically be acquired through literature review and field investigations. Transforming these into fixed procedures requires only the application of software engineering principles. However, in reality, the environment is invariably dynamic, and pre-written programs in Python or C++ are incapable of handling constantly changing boundary conditions. The primary limitation of traditional programming lies in its tendency to produce uncontrollable outputs when inputs fall outside predefined parameters. In contrast, while the functionalities and specific calculations provided by the large language model-driven Navigation-GPT framework—such as encounter prediction and risk analysis—are not significantly different from those of conventional intelligent systems, the core distinction lies in its ability to respond to environmental anomalies. When inputs deviate from preset conditions, the new LLM-based framework does not require program modifications but instead derives relatively reasonable solutions through intrinsic relational analysis. This capability makes it feasible to develop an intelligent system capable of addressing most navigational environments.

The comparative experiments between Navigation-GPT and DeepSeek-R1-Distill-Qwen-32B (DeepSeek), GPT-4o indicate that, in the field of collision avoidance, the performance of the LoRA-fine-tuned, smaller 7B model significantly outperforms the 32B model distilled by DeepSeek-R1 and GPT-4o. After deploying the same framework of Navigation-GPT, especially integrating external tools such as DCPA and TCPA, the performances of the smaller fine-tuned LLM core, DeepSeek and GPT-4o becomes similar. However, the smaller fine-tuned LLM core exhibits shorter response delays, with a decision-making time of approximately 15 seconds, while DeepSeek requires about 58 seconds. After fine-tuning, the smaller LLM core generates recommendations that are more aligned with navigational terminology, more accurate, and free from model hallucinations.

It is noteworthy that the computational capabilities of LLMs are inherently limited. Consequently, complex risk parameters computed solely by LLMs may lack reliability. To address this limitation, integrating external tools to expand the action space of LLMs and construct agents remains the primary technical approach for solving complex tasks involving intricate computations. This hybrid methodology leverages the strengths of both LLMs and specialized tools, thereby enhancing overall system performance and robustness.

Unlike other vertical applications of large models, Navigation-GPT adopts a dual-large model kernel architecture. One kernel is responsible for tool invocation, while the other is in charge of information verification and conclusion formation. This design not only ensures the integrity of information collection but also effectively reduces the risk of model hallucinations. Navigation-GPT demonstrates functionality similar to traditional collision avoidance algorithms. However, compared to conventional hard-coded intelligent systems, the unique open-scene processing capability of Navigation-GPT provides a new framework for the development of intelligent navigation systems.

\section{Conclusion}
This research proposed Navigation-GPT, a flexible framework designed to adapt to open navigation scenarios. Navigation-GPT integrates technologies such as ReAct and LoRA to build an LLM-based agent capable of processing heterogeneous data from onboard sensors to perform navigation and collision avoidance tasks. Extensive experiments demonstrate that Navigation-GPT functions similarly to traditional intelligent systems but can handle unstructured scenarios that traditional systems cannot manage. Furthermore, a comparison experiment with the latest DeepSeek-R1 indicates that the LoRA module integrated into Navigation-GPT enhances the performance of smaller LLMs in specialized domains. An ablation experiment on DeepSeek-R1 shows that current SOTA LLMs cannot be directly applied to navigation directly, and using external computational tools to assist agent decision-making remains the primary approach for LLMs in executing complex tasks. Since Navigation-GPT currently focuses on navigation and collision avoidance, it lacks long-distance path planning capabilities. Future research will further explore the application of LLMs in ship path planning to improve the robustness of Navigation-GPT.

\textbf{Declaration of competing interest}

The authors declare that they have no known competing financial interests or personal relationships that could have appeared to influence the work reported in this paper.

\textbf{Acknowledgments}

This work is financially supported by the Funds for the National Key R$\&$D Program of China (Grant No. 2023YFB4302300), National Natural Science Foundation of China under Grant No. 52201415, Research on the Implementation Plan of the New Generation Shipping System in Pingdingshan, Luohe, and Zhoukou, Fund of State Key Laboratory of Maritime Technology and Safety (Grant No. 16-10-1).

\bibliographystyle{elsarticle-num} 
\bibliography{ref.bib}

\end{document}